\documentclass[sigconf, 10pt]{acmart}

\usepackage[english]{babel}
\usepackage{blindtext}

\renewcommand\footnotetextcopyrightpermission[1]{} 
\setcopyright{none}

\acmDOI{}

\acmISBN{}

\acmPrice{}

\usepackage[utf8]{inputenc} 
\usepackage[T1]{fontenc}    
\usepackage{hyperref}       
\usepackage{url}            
\usepackage{booktabs}       
\usepackage{amsfonts}       
\usepackage{nicefrac}       
\usepackage{microtype}      
\usepackage{xcolor}         
\usepackage{graphicx,wrapfig}

\usepackage{amsmath}
\usepackage{mathtools}
\usepackage{amsthm}
\usepackage{algorithm}
\usepackage{algorithmic}
\usepackage[caption=false]{subfig}

\usepackage{diagbox}
\usepackage{tikz}
\usepackage{xurl}
\usepackage{pgfplots}
\usepackage{tabularx,booktabs}
\usetikzlibrary{plotmarks}
\usetikzlibrary{patterns}
\usepackage{mdframed}
\definecolor{mygray}{RGB}{211,211,211} 
\usepackage{pgfplotstable}
\pgfplotsset{compat=1.7}
\usepgfplotslibrary{groupplots}
\newlength\myindent
\usepackage{xspace}

\newcommand{\fbpage}{\textmd{Facebook}\xspace}
\newcommand{\lastfm}{\textmd{LastFMAsia}\xspace}
\newcommand{\cora}{\textmd{Cora}\xspace}
\newcommand{\seer}{\textmd{Citeseer}\xspace}
\newcommand{\pubmed}{\textmd{PubMed}\xspace}

\newcommand{\reddit}{\textmd{Reddit}\xspace}
\newcommand{\products}{\textmd{Products}\xspace}

\newcommand{\Gc}{\mathcal{G}}
\newcommand{\Vc}{\mathcal{V}}
\newcommand{\Ec}{\mathcal{E}}

\newcommand{\numlabels}{O}
\newcommand{\numrounds}{R}
\newcommand{\hopiter}{m}

\newcommand{\feat}{\mathbf{F}}

\newcommand{\num}{K\xspace}

\newcommand{\fc}[1]{\mathbf{#1}} 
\newcommand{\cf}[1]{\mathsf{#1}} 
\newcommand{\arr}[1]{\mathbf{#1}} 
\newcommand{\set}[1]{\mathcal{#1}} 

\newcommand{\tool}{\textsc{Retexo}\xspace}
\newcommand{\toolgnn}{\textsc{RetexoGNN}\xspace}
\newcommand{\toolgnns}{\textsc{RetexoGNN}s\xspace}

\newcommand{\toolgcn}{\textsc{RetexoGCN}\xspace}
\newcommand{\toolsage}{\textsc{RetexoSage}\xspace}
\newcommand{\toolgat}{\textsc{RetexoGAT}\xspace}
\newcommand{\sage}{GraphSAGE\xspace}
\newcommand{\gcn}{GCN\xspace}
\newcommand{\gat}{GAT\xspace}

\newcommand{\sagea}{BNS-SAGE (0.1)\xspace}
\newcommand{\gcna}{BNS-GCN (0.1)\xspace}
\newcommand{\gata}{BNS-GAT (0.1)\xspace}

\newcommand{\gatb}{BNS-GAT (0.01)\xspace}

\definecolor{amethyst}{rgb}{0.6, 0.4, 0.8}
\definecolor{azure(colorwheel)}{rgb}{0.0, 0.5, 1.0}
\definecolor{chocolate(traditional)}{rgb}{0.48, 0.25, 0.0}
\definecolor{onyx}{rgb}{0.06, 0.06, 0.06}

\pgfplotscreateplotcyclelist{fycle}{
    {red},
    {blue}, 
    {amethyst},
    {azure(colorwheel)}
    {orange},
    {chocolate(traditional)},
}

\pgfplotscreateplotcyclelist{exotic}{%
magenta,every mark/.append style={fill=magenta!80!black},mark=*\\%
brown,every mark/.append style={fill=brown!80!black},mark=square*\\%
blue!60!black,every mark/.append style={fill=blue!80!black},mark=otimes*\\%
red!70!white,mark=star\\%
amethyst!80!black,every mark/.append style={fill=amethyst},mark=diamond*\\%
red,densely dashed,every mark/.append style={solid,fill=red!80!black},mark=*\\%
yellow!60!black,densely dashed,
every mark/.append style={solid,fill=yellow!80!black},mark=square*\\%
black,every mark/.append style={solid,fill=gray},mark=otimes*\\%
blue,densely dashed,mark=star,every mark/.append style=solid\\%
red,densely dashed,every mark/.append style={solid,fill=red!80!black},mark=diamond*\\%
}

\pgfplotscreateplotcyclelist{exotic1}{%
magenta,every mark/.append style={fill=magenta!80!black},mark=none,mark size=3pt\\%
black,every mark/.append style={fill=black!80!black},mark=none,mark size=4pt\\%
blue!60!black,every mark/.append style={fill=blue!80!black},mark=none,mark size=3pt\\%
red!70!white,mark=none\\%
amethyst!80!black,every mark/.append style={fill=amethyst},mark=none\\%
red,densely dashed,every mark/.append style={solid,fill=red!80!black},mark=none\\%
yellow!60!black,densely dashed,
every mark/.append style={solid,fill=yellow!80!black},mark=none\\%
black,every mark/.append style={solid,fill=gray},mark=otimes*\\%
blue,densely dashed,mark=star,every mark/.append style=solid\\%
red,densely dashed,every mark/.append style={solid,fill=red!80!black},mark=diamond*\\%
}

\ExplSyntaxOn
\DeclareExpandableDocumentCommand \round { O{0} m }
 { \fp_eval:n { round(#2,#1) } }
\ExplSyntaxOff

\begin{document}

\title{Scalable Neural Network Training \\ over Distributed Graphs}

\author{
Aashish Kolluri,\textsuperscript{*} Sarthak Choudhary,\textsuperscript{*} Bryan Hooi, and Prateek Saxena\\
\text{\Large School of Computing, National University of Singapore}\\
\text{\Large aashish7@comp.nus.edu.sg, csarthak76@gmail.com, \{bhooi, prateeks\}@comp.nus.edu.sg}
}

\begin{abstract}
Graph neural networks (GNNs) fuel diverse machine learning tasks involving graph-structured data, ranging from predicting protein structures to serving personalized recommendations. Real-world graph data must often be stored distributed across many machines not just because of capacity constraints, but because of compliance with data residency or privacy laws. In such setups, network communication is costly and becomes the main bottleneck to train GNNs. Optimizations for distributed GNN training have targeted data-level improvements so far---via caching, network-aware partitioning, and sub-sampling---that work for data center-like setups where graph data is accessible to a single entity and data transfer costs are ignored.

We present \tool, the first framework  which eliminates the severe communication bottleneck in distributed  GNN training while respecting any given data partitioning configuration. The key is a new training procedure, {\em lazy message passing}, that reorders the sequence of training GNN elements. \tool achieves {\em $1-2$ orders of magnitude reduction} in network data costs compared to standard GNN training, while retaining accuracy. \tool scales gracefully with increasing decentralization and decreasing bandwidth. It is the first framework that can be used to train GNNs at all network decentralization levels---including centralized data-center networks, wide area networks, proximity networks, and edge networks.
\end{abstract}

\maketitle
\pagestyle{plain}

\begingroup\renewcommand\thefootnote{*}
\footnotetext{These authors contributed equally to this work.}
\endgroup
\section{Introduction}
\label{sec:intro}

Machine learning on graphs is a fundamental problem, with classical solutions such as cluster analyses and label propagation (see survey~\cite{fortunato2010community}). Graph neural networks (GNN) have surpassed prior machine learning solutions by offering state-of-the-art performance in many machine learning tasks over graph data such as serving recommendations, financial fraud detection, and drug discovery~\cite{wu2020comprehensive,zhou2020graph}. Often in these applications, the graphs can be very large, therefore systems that optimize the cost of training GNNs are on the rise~\cite{wan2022bns,gandhi2021p3,zheng2022bytegnn}.

We are in an era where data sovereignty, geographic data residency, privacy laws like GDPR, and intellectual property concerns play a pivotal role in governing the use of sensitive information. Such governance applies to graph data used in machine learning systems as well. It is often necessary to {\em partition} such data to ensure that sensitive user data and processing is confined solely at certain servers or client devices~\cite{mcmahan2017communication,kairouz2021advances}. This is to legally comply with data use policies and to manage the risk of data breaches~\cite{metafine}. Such partitioning requirements are often externally imposed, e.g., no single entity may store the entire dataset, and regulations forbid its exchange without user consent. Performance optimization must respect such data decentralization constraints.

Several works have been proposed to optimize training GNNs on partitioned graph data across networked machines (workers)~\cite{wan2022bns,  md2021distgnn, gandhi2021p3, zheng2022bytegnn,zhu2019aligraph,cai2021dgcl,jia2020improving,zheng2020distdgl,liu2023bgl}. These works have established that {\em data-intensive message passing} rounds conducted over the network between workers is the major bottleneck while training GNNs on large partitioned graphs. Our experiments presented later confirm the same: With the standard training procedure, over $2$ TB of data is communicated over the network for end-to-end training on a graph with close to $2.5$ million nodes when partitioned among just $2$ workers. Therefore, reducing the network data volume (costs) to train GNNs is critical for many reasons---lowering financial costs, reducing training time, and democratizing training even over low-bandwidth or unreliable networks.

Prior works, while highlighting the bottleneck, propose solutions for the centralized training setups where the entire graph is accessible to a centralized coordinator and worker machines can access each other's raw data arbitrarily over fast communication links. Their systems implement variants of mainly three data-level optimizations: graph- and network-aware partitioning, caching data from other workers, and graph sub-sampling. None of them are designed for general distributed setups with arbitrary data decentralization constraints, and for handling diverse network characteristics, as a first-class principle. For instance, no single entity partitioning the graph is often feasible when graph data resides at workers separated by geopolitical boundaries. In centralized training setups, data transfer over the internal network does not incur much financial costs. Further, bandwidth constraints can also be eased with specialized infrastructure, for instance, RDMA networks with a throughput of over $100$ Gbps~\cite{liu2023janus}. Network data transfer costs and the available bandwidth is still a critical bottleneck for all other setups where workers may be GPUs connected over NICs / PCIe interfaces or devices connected over wide-area networks, via edge servers, or over wireless/bluetooth. These settings may have less than $10$ Gbps links between workers and may require thousands of dollars to train a single medium-size GNN (see Section~\ref{sec:problemndc}).

\paragraph{Our Solution: \tool.}
We design the first execution framework, called \tool, for training GNNs efficiently for all levels of graph data decentralization. \tool rethinks the training process of GNNs from the perspective of communication-efficiency, i.e., reduce as much network data volume as possible. 
The key idea is to change the {\em order in which layers of the neural network} are trained giving an effect of delaying the message passing operations during training until necessary. We call it the {\em lazy message passing} training strategy.
Figure~\ref{fig:finer_comm_graphsage_2_fed} in our evaluation illustrates \tool's savings of $1$ to $2$ orders of magnitude in network data costs, compared to state-of-the-art~\cite{wan2022bns} and standard training strategies respectively. 


\tool is designed to adhere to any given data partitioning requirements, i.e., \tool does not need to share raw node features or graph edges with any centralized coordinator. Communication can be peered between workers, so cross-worker graph edges need not be routed via a centralized coordinator. This design, first and foremost, {\em removes any algorithmic necessity to route data centrally}. The centralized coordinator is only used to conduct the training process in a synchronized manner and aggregate the gradients sent by the workers for every training round. 
Minimizing centralized coordination has well-known benefits of avoiding performance bottlenecks, but it also gives better data controls---each worker has complete autonomy over  what it sends to other workers and the coordinator. It offers a natural baseline to implement advanced privacy enhancements in the future, for instance, local differential privacy with noise tailored individually to each worker~\cite{sajadmanesh2021locally,zhu2023blink,kolluri2022lpgnet}.

\tool can achieve up to $321\times$ better network data costs compared to standard GNN training on standard benchmarks. It has $32 \times$ better costs compared to BNS-GCN, the state-of-the-art system designed to minimize cross-worker costs. As a consequence, \tool also trains GNNs end-to-end up to $22\times$ and $2.7\times$ faster than standard GNN training and BNS-GCN respectively, on our testbed which is similar to prior works~\cite{wan2022bns,gandhi2021p3}. The accuracy offered by \tool for these architectures is within $\pm1\%$ of that when centrally trained. At the same time, it respects data partitioning constraints, unlike prior systems~\cite{gandhi2021p3} which may violate these constraints by replicating raw data across workers.

\paragraph{Compatibility with GNNs and Training Frameworks.} \tool is compatible with any message passing GNN model architectures such as GCN~\cite{kipf2016semi}, GraphSAGE~\cite{hamilton2017inductive}, and GAT~\cite{velivckovic2017graph}.  We have implemented \tool on two popular distributed machine learning frameworks---PyTorch Distributed~\footnote{ \url{https://pytorch.org/tutorials/beginner/dist_overview.html}} and Flower~\cite{beutel2020flower}. This illustrates that \tool can work in two very different network setups: multi-machine GPU clusters and bandwidth-constrained mobile Raspberry Pi (RPi) clients.
To the best of our knowledge, \tool is the first practical framework to enable training GNNs on distributed graphs stored on mobile clients. We have released the code to reproduce our experiments.\footnote{https://github.com/aashishkolluri/retexo-distributed} We believe this work does not raise any ethical issues.

\section{Background \& Overview}
\label{sec:problem}

Training GNNs on multiple workers distributively is indispensable for large datasets found in social networks~\cite{ying2018graph}, financial transactions~\cite{awsaifinancial,amazonscalable}, and recommender systems~\cite{linkedin} due to insufficient compute and memory in a single worker.

Another practical reason is compliance with data residency and privacy regulations. Real-world graph data often originates from distributed sources. For example,
client users often generate data that are part of graphs arising in social networking or web analytics. These graphs can be sensitive (e.g. signifying personal relationships or consumer preferences)~\cite{korolova2008link,wondracek2010practical}. It is desirable to minimize the amount of sensitive graph data shared externally to the device for privacy reasons. Similarly, data residing at various edge servers, geo-distributed data centers, and across different organizations is often required to stay within jurisdictional boundaries and subject to cross-border transfer regulations otherwise~\cite{facebookdataresidency,incountryblog,awsdataresidency}. Therefore, a good system for distributed GNN training should not assume that the data placement can be arbitrary, rather it should work with any data partitioning imposed by external constraints. The general edge-partitioned setup we describe next accommodates both motivations: efficiency and data decentralization.



\subsection{Edge-partitioned Distributed Setup}

\begin{figure}[t]
    \centering
    \includegraphics[scale=0.35]{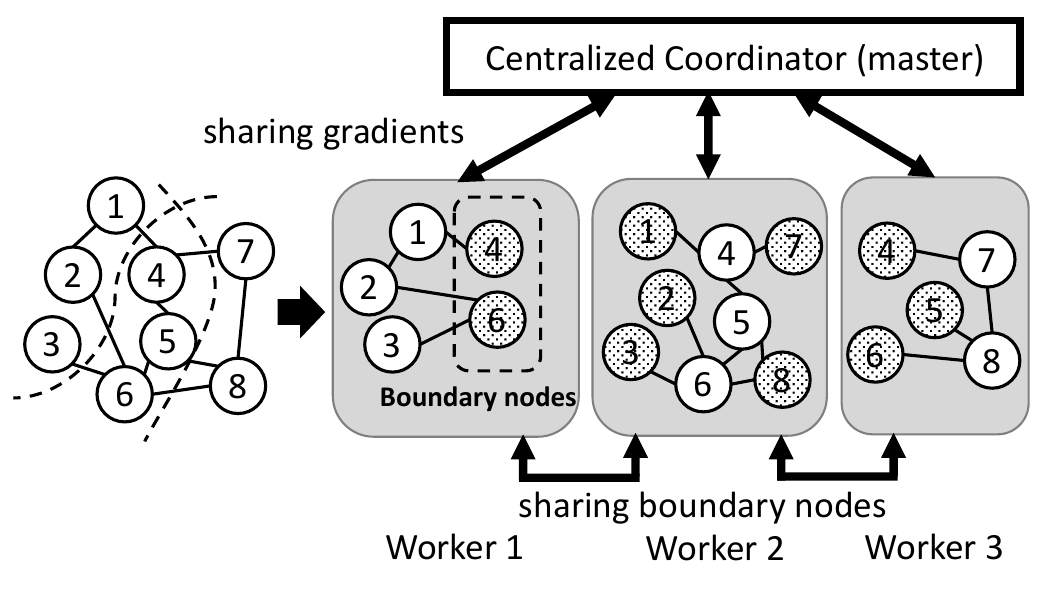}
    \caption{Edge-partitioned Distributed Setup}
    \label{fig:fedsetup}
\end{figure}

Consider a graph $\Gc:(\Vc, \Ec)$, where $\Vc$ and $\Ec$ are the set of nodes and edges respectively.
The graph $\Gc$ is partitioned across $N$ workers, i.e., each worker holds a set of {\em inner nodes} $\Vc_i \subset \Vc$ it owns and all the edges between its inner nodes. A {\em boundary node} for a worker $i$ is an inner node from another worker that has an edge with an inner node at $i$. Therefore, each worker $i$ also stores its boundary nodes and the boundary edges i.e., the edges between its inner nodes and its boundary nodes. We want the edge-partitioned distributed setup to be general enough to address any level of partitioning, from completely centralized to fully decentralized. Therefore, we assume that these partitions exist naturally i.e., no entity controls what the partitions can be. Further, none of the workers share their graph data with other workers apart from the boundary edges which the corresponding workers are mutually aware of. The graphs are undirected in our work, though one can extend to directed graphs in a straightforward way.  We illustrate this setup in Figure~\ref{fig:fedsetup}.

We want to train GNNs over such graphs {\em efficiently}.
Our focus is on the most common application, namely supervised node classification. All nodes have {\em feature vectors} i.e., each node in $\Vc$ has a feature vector of $I$ dimensions captured by the node-feature map $\feat:\Vc\xrightarrow{}\mathbb{R}^{I}$. For example, in a social networking graph, the feature vectors can be word2vec embeddings~\cite{mikolov2013efficient} of user post topics. A few of the nodes have {\em labels}, for instance, topics that a user (node) in a social network is interested in.
The task is to train a GNN on the graph to classify other nodes that do not have labels. 

The features and other intermediate representations of a node have to be shared with its neighbors during the training of a GNN. It is therefore necessary for every worker to access the features and intermediate representations of their boundary nodes from other workers. But, nothing else is shared across workers, keeping with the {\em principle of least privilege}.


Our setup has a centralized coordinator (master) which collaborates with other workers for training. A worker denotes a process utilizing the resources of a machine with either a single or multiple GPU/CPU. In practice, multiple workers could be deployed on a single machine. To visualize the problem better, one can assume that a unique worker is deployed per machine that has a single GPU/CPU, communicating with other workers via network interfaces. The master only schedules training rounds and helps in aggregating gradient vectors from the workers~\cite{mcmahan2017communication}. The master may also assist the workers in setting up trusted peer-to-peer communication channels among them at the beginning if they do not exist already. 
During the entire training, the master will not be able to access raw node features or the graph structure from any worker and the workers are oblivious to any features, intermediate representations, or edges stored on other workers except the representations of their boundary nodes. Finally, we point out that this setup echoes the federated learning setup for images and text that is primarily motivated from a data decentralization perspective~\cite{mcmahan2017communication}.


\subsection{Architecture of Classical GNNs} 
\label{sec:backgroundgnnarch}

\begin{figure}[t]
    \centering
    \includegraphics[scale=0.47]{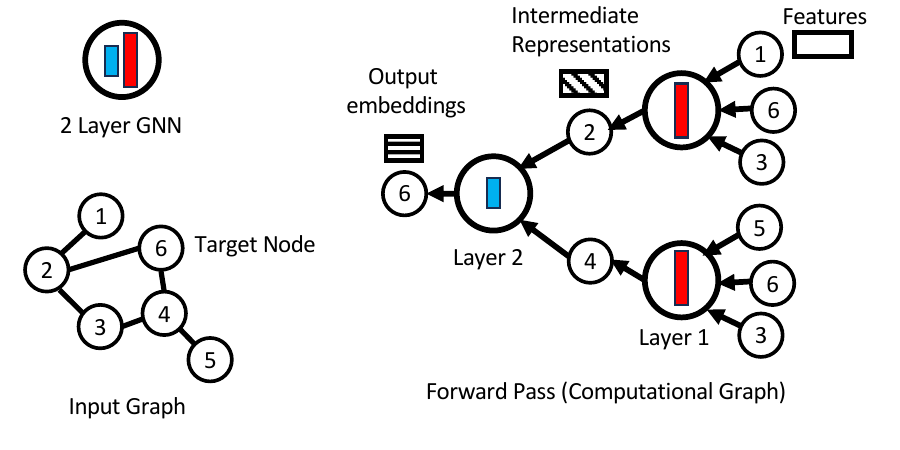}
    \caption{GNN model architecture}
    \label{fig:gnn_arch}
\end{figure}

GNNs are feedforward neural networks. The inputs and outputs of every intermediate layer in the neural network is a vector often called a {\em representation}. We focus on {\em message passing} GNNs which are widely-used in practice, such as the GCN~\cite{kipf2016semi}, GraphSAGE~\cite{hamilton2017inductive}, and GAT~\cite{velivckovic2017graph}. 
We illustrate the forward pass of a $2$-layer GNN to compute the embeddings for a node in Figure~\ref{fig:gnn_arch}.
During the forward pass, every node aggregates representations from its neighbors and combines them with its own representations in each layer $\hopiter \in [1,\ldots,\num]$. Each GNN layer can be viewed as performing one message passing round. After $m$ layers (or message passing rounds), a node's representation has aggregated information from all nodes in its $m$-hop neighborhood. 

Formally, the GNN model architecture can be viewed as the function $\fc{M}:(v,\feat, \Ec))\xrightarrow{}\mathbb{R}^\numlabels$. $\fc{M}$ takes a node $v$, all features of all nodes $\feat$ (which includes the $m$-hop neighborhood of $v$), and the edges $\Ec$ as input. It outputs the node embedding vectors (representations) of $\numlabels$ dimensions for the given node $v$. 
The raw features of $v$ are the representations used as input to the first GNN layer.
%
Therefore, the representation of the node $v$ after $\hopiter$ rounds of message passing will be:
{\small 
\begin{align}
    \arr{Q}^{\hopiter}_v = \fc{f}^{\hopiter-1}_\theta(\cf{COMBINE}(\arr{Q}^{\hopiter-1}_v, \cf{AGGR}^{\hopiter-1}_{u\in\set{N}(v)}(\arr{Q}^{\hopiter-1}_u)))\text{,} \arr{Q}^0_u = \feat[u]  \label{eq:1}
\end{align}
}%

Here, the neighbors of $v$ are denoted by $\set{N}(v)$, the\\ $\cf{COMBINE}$ function is usually vector concatenation, the non-linear functions $\fc{f}^{\hopiter-1}_\theta$ and $\cf{AGGR}$ may have trainable parameters. $\cf{AGGR}$ functions vary with different GNNs. For example, $\cf{AGGR}$ is the mean operator in GCNs~\cite{kipf2016semi} or a max-pooling layer in \sage~\cite{hamilton2017inductive}. $\fc{f}^{\hopiter-1}_\theta$ represents a sequence of non-linear functions including a fully-connected layer, the ReLU function, batch, and layer normalization layers. The $\hopiter^{th}$ {\em GNN layer} can be represented in short using only the functions with trainable parameters $(\cf{AGGR}^{\hopiter-1}, \fc{f}^{\hopiter-1}_\theta)$. The $\hopiter^{th}$ representation of a node is, thus, a non-linear transformation over an aggregate of the $({\hopiter}-{1})^{th}$ neighbors' representations. 

The last layer produces an output embedding, on which labels are computed for the classification task.
%
%
%
The goal of training is to optimize the values of $\fc{f}^{\hopiter-1}_\theta$ and $\cf{AGGR}^{\hopiter-1}$ functions for all $m \in\{1,\ldots,\num\}$ for high classification accuracy.

\subsection{Problem: Network Data Costs}
\label{sec:problemndc}

Training GNNs distributively can incur massive network costs due to their inherent message passing architecture.

\paragraph{Standard Training.} The usual procedure to train GNNs distributively proceeds in {\em training rounds}.  A training round has two phases: local training and gradient aggregation~\cite{wan2022bns,thorpe21dorylus,jia2020improving}. 
The workers receive the same parameters of model $\fc{M}$ at the start of each training round. The local training uses the standard stochastic gradient descent algorithm~\cite{bottou2018optimization}, which executes a forward pass execution of $\fc{M}$ to compute the loss function and then a backward pass of $\fc{M}$ to compute gradient vectors. During the forward pass, the workers send and receive the intermediate representations of boundary nodes with other workers, once for each layer in the GNN. During the backward pass, each worker receives the gradients corresponding to the representations they sent during the forward pass for a boundary node. The worker aggregates the received gradients for each layer for each node, working backward from the layer $K$ to $1$. Thus, there are $2$ message passing rounds per layer and $2\cdot\num$ in total for a $\num$-layer GNN. 

\begin{figure*}[t]
    \centering
    \includegraphics[scale=0.4]{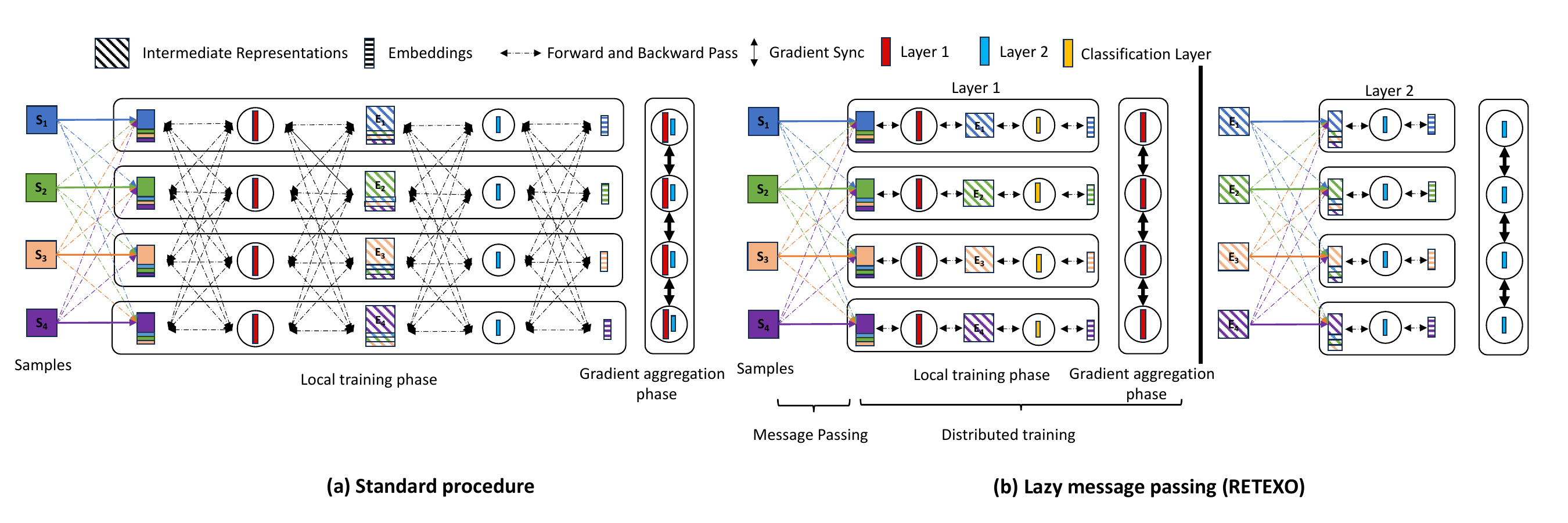}
    \caption{Comparison of training using standard procedure (left) and Lazy message passing (right).}
    \label{fig:baseline_and_retexo}
\end{figure*}

At the end of a local training phase, each worker has a local gradient computed for all trainable parameters of the GNN. All workers send their local gradients to the master, which aggregates them, updates $\fc{M}$, and sends the updated $\fc{M}$ for the next training round. This constitutes the gradient aggregation phase, completing one training round. Figure~\ref{fig:baseline_and_retexo} illustrates the standard training for one end-to-end round.

\begin{figure}[t]
    \begin{tikzpicture}
        \begin{groupplot}[group style = {group size = 1 by 1, horizontal sep = 100pt, vertical sep = 35pt, xlabels at=edge bottom, ylabels at=edge bottom},
                width=0.73*\columnwidth,
                height = 3.5cm,
                x label style={font=\small},
                y label style={font=\small},
                ylabel={seconds},
                ]
            \input{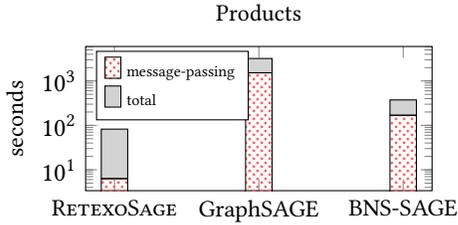}
        \end{groupplot}
    \end{tikzpicture}
    \caption{Time taken for end-to-end training (log scale).}
    \label{fig:training_time_sage}
\end{figure}

\paragraph{Network data costs with an example.}
It is easy to see that the network data volume exchanged between workers in the standard training increases linearly with the number of boundary nodes. Additionally, it increases linearly with the number of training rounds. If one partitions the graph randomly among a few workers then the number of boundary nodes can be in millions for even medium-sized graphs, such as the Amazon \products co-purchasing graph which has $2.5 \times 10^{6}$ nodes~\cite{hu2020open}. In the standard benchmarked settings, the network data volume is close to $2.8$ GB just to share one intermediate representation among just $2$ workers for this graph.  Further, close to $3.7$ TB of data volume is shared over the network among the workers for training a small GNN with $3$ layers on the Amazon \products graph. In the centralized setup itself, such high network data volume is prohibitive as pointed out by prior works~\cite{gandhi2021p3,wan2022bns}, specifically in the context of reducing training time. This problem becomes even worse for the edge-partitioned distributed setup due to the network traffic being routed over bandwidth-constrained and unreliable communication channels, incurring high costs of such data transfer.

As an example, consider two workers hosted over Amazon Web Services (AWS) which are located in two different continents. The costs of sending data even within AWS services can be up to $0.1$ USD per GB of data. Therefore, to train a GNN once for our running example, one would end up spending $370$ USD and about $1$ USD for every inference on the GNN. Extrapolating from this example, the cost to train a GNN on another commonly used benchmark, ogbn-papers100M ~\cite{hu2020open}, with $111$ million nodes is $17,000$ USD and the cost of one inference is $43$ USD. We point out that GNNs are trained on these datasets regularly in the centralized setup nearly for free.\footnote{Leaderboard: \url{https://ogb.stanford.edu/docs/leader_nodeprop/}} Such network costs are less relevant in centralized training setups, where data is accessible by GPU workers connected over extremely fast links (e.g. NvLinks) on the same machine; here, computational costs dominate~\cite{liu2023janus}.

Further, high network data volume inevitably affects the training time. Figure~\ref{fig:training_time_sage} illustrates the training time to train GraphSAGE on \products dataset using the standard strategy (middle bar). Observe that more than $80\%$ of the training time is spent during the message passing rounds. This will worsen as the bandwidth reduces in the distributed setups. The remaining bars provide a glimpse of training time using \tool (first) and another state-of-the-art framework~\cite{wan2022bns}.


\subsection{Prior Optimization Approaches}
\label{sec:prioropt}

Many prior works have proposed optimization techniques that reduce network data costs of training GNNs on distributed graphs~\cite{gandhi2021p3,wan2022bns,zhu2019aligraph,zheng2022bytegnn,cai2021dgcl,zheng2020distdgl}. However, they assume that a central coordinator is in charge of optimization can customize the data partitioning, as needed, to reduce cross-worker costs. Two prominent examples of such optimizers are graph structure based~\cite{zheng2020distdgl,zheng2022bytegnn,cai2021dgcl} and node feature partitioning~\cite{gandhi2021p3}
which optimizes data placement for efficiency. Many of these works also assume direct access to features and edges of inner nodes, beyond boundary nodes, residing at other workers. These are {\em incompatible} with regulatory goals: Adhering to any partitioning pre-decided by the setup and disabling direct data access of inner nodes is desirable.


One and the only prior optimization approach that fits our constraints is boundary node sampling (BNS)~\cite{wan2022bns}. 
The idea is that a worker samples a small fraction of its boundary nodes at another worker, for which it exchanges data during each training round. This reduces communication costs proportional to the sampling rate.
However, if the sampling rate is chosen too aggressively (say below $10\%$), the model performance is adversely affected in each round, and the number of training rounds needed for the GNN model to converge increases, as pointed out in the original work that proposed BNS~\cite{wan2022bns}. 
Despite being an orthogonal approach to ours, it does offer a good experimental comparison point, since the BNS framework has implemented sampling as well as other optimizations to reduce network data costs proposed in prior work, including caching boundary node features, precomputing inputs to specific GNNs for all training rounds, and pipelining training with evaluation.

\section{Training GNNs with \tool}

Recall that the network data cost to train a GNN increases linearly with the number of boundary nodes and message passing rounds. All existing techniques focus on reducing the network data cost of each message passing round by using optimal partitioning. We observe that it is insufficient to reduce the network data cost of individual message passing rounds. Even a single local training phase with $2\cdot\num$ message passing rounds may generate more network data than that generated during the gradient aggregation phase across all training rounds combined (see Section~\ref{sec:commeffl}). Furthermore, there would be $2\cdot \num \cdot R$ such message passing rounds to train a $\num$-layer GNN end-to-end for $R$ training rounds using standard training. Therefore, {\em our key idea is to reduce the number of message passing rounds itself}, while still having enough of them to achieve high model performance. To do this, we propose a novel training procedure for GNNs without changing their underlying architecture. It embodies what we call the {\em lazy message passing} training strategy.

\subsection{Lazy Message Passing}
\label{sec:lmp}

We propose to train GNNs layer-by-layer 
instead of optimizing the parameters of all layers together. Specifically, each GNN layer is trained independently of the other layers and sequentially after training all of its previous layers. The rationale is as follows. Observe that while using a trained $\num$-layer GNN for inference, it can perform the classification task using the $\num$-hop neighborhoods of all the nodes well.
This means that once the $\num$ layers are trained, the output representations of $\num^{th}$ layer are informative.
Furthermore, the total number of message passing rounds is just $\num$ during inference since only one forward pass is needed. 

We can also use the above insight to make training efficient. Suppose we have already trained the parameters of the $K$ layers and fix them thereafter. Now, if we were to add one more layer, say a $\{\num+1\}^{th}$ layer, we can use the representations already learned well by the trained $\num$-layers as inputs to the untrained final layer. To train this single $\{\num+1\}^{th}$ layer, the workers perform one round of message passing to form the representations needed as inputs to the $\{\num+1\}^{th}$ layer. Then, each worker can locally train this layer {\em with no more message passing rounds}, since all the information to run the forward and backward passes is then available locally. 



We can use this insight to train the first layer itself, and beyond. The inputs to the first layer are obtained after one message passing round over raw features. After the first layer is trained the subsequent layers can be trained by using the representations obtained from the last trained layer as inputs. This has the effect of delaying a message passing round until completely training all previous layers of the GNN. Hence, we call this greedy layer-by-layer optimization procedure to train GNNs as {\em Lazy Message Passing}.

\paragraph{Concrete Illustration.}   Consider a $2$-layer GNN with layers ($\cf{AGGR^0}$, $\fc{f}^0_\theta$) and ($\cf{AGGR^1}$, $\fc{f}^1_\theta$) following the definition in equation~\ref{eq:1} from Section~\ref{sec:backgroundgnnarch}. Before training the first layer  one round of message passing is performed to update the feature information available at every node with their neighbors'. Using the updated feature information the first layer is trained independently from the second layer. To do that a classification layer, i.e., a non-linear function, $\fc{C}^0_\theta$ is added to the first layer. Effectively, the layer ($\cf{AGGR^0}$, $\fc{f}^0_\theta$, $\fc{C}^0_\theta$) is trained to perform the node-classification task by itself. Once trained, the classifier layer $\fc{C}^0_\theta$ is removed, hence, leaving behind the first layer that has learned to aggregate information from $1$-hop neighbors in a way that is useful for node classification. Observe that training the first layer did not require any more message passing rounds since there is only one $\cf{AGGR}$ function in that layer. Now, the second GNN layer ($\cf{AGGR^1}$, $\fc{f}^1_\theta$) is trained. The inputs to this layer are the fixed intermediate representations output by the trained first layer, i.e., $\arr{Q^1}$. At this point, another message passing round is conducted for every node to obtain its neighbor's intermediate representations. Thus, each node now has information from its $2$-hop neighborhood. Using this information, the second GNN layer is trained similarly to the first one.  We do not need to add a classification layer here because $\fc{f}^1_\theta$ is already the final classification function of a $2$-layer GNN. Further, the first layer's weights are not updated while training the second layer. Following this process, we have trained a GNN to do node classification while only performing $2$ rounds of message passing without changing its architecture.

We contrast our procedure to the standard training shown in Figure~\ref{fig:baseline_and_retexo}. The standard training tries to optimize parameters of all layers jointly, and therefore, incurs $2\cdot \num$ every training round. Midway during training, when the first $K$ layers are only partly optimized, i.e., have poor classification
power, the subsequent layers receive noisy input representations and the final computed loss is high. It takes several training rounds to reduce the loss in standard training.
The message passing incurred in each training round is excessive. 

Lazy message passing avoids these expensive costs by improving the $\num$-hop classification accuracy for nodes before  proceeding to improve it over the $(\num+1)$-hop.
We will show in our evaluation (Section~\ref{sec:eval_accuracy}) that the node classification performance of our strategy is as high as the standard training in several benchmarked graph datasets.

\begin{table}[t]
    \centering
    \caption{Summary of symbols used, size in bytes.}
    \resizebox{\columnwidth}{!}{
    \begin{tabular}{c|c}
        symbol & description \\\hline
        $R$ & \# training rounds\\
        $n$ & \# workers\\
        $B$ & \# boundary nodes per worker\\
        $H_0$ & input features size\\
        $H_m$, $m \in [1\ldots \num-1]$ & intermediate representations size \\
        $C_m$ & size of additional classifier layer\\
        $M$ & size of the GNN \\\hline
    \end{tabular}
    }
    \label{tab:symbols}
\end{table}

\subsection{Communication-efficiency Analysis}
\label{sec:commanalysismain}
It is illustrative to compare the total data costs analytically for lazy message passing with the standard procedure. During any particular phase of training, we denote the network data costs as the size of the data that has been communicated over the network by all the workers during that phase. We also call it the network data volume. We present the main equations for the analytical network data volume here and provide detailed explanations in Appendix~\ref{appdx:commanalysis}.

Considering both local training and gradient aggregation phases, the ratio of the total network data volumes for standard ($DV^{standard}_{total}$) and lazy message passing ($DV^{lazy}_{total}$) strategies, which we define as $\gamma$ comes out to be,

\begin{align}
    \gamma = \frac{DV^{standard}_{total}}{DV^{lazy}_{total}} = \frac{1 + B\cdot \left(2\frac{\sum_{\hopiter=1}^{\num-1}H_{\hopiter}}{M} + \frac{H_0}{M\cdot R}\right)}{1 + \frac{\sum_{\hopiter=1}^{\num-1}{C_{\hopiter}}}{M} + B\cdot\left(\frac{\sum_{\hopiter=1}^{\num-1}H_{\hopiter}}{M\cdot R} + \frac{H_0}{M\cdot R}\right)} \label{eq:2}
\end{align}

Consider a scenario where network data volume during one round of local training is higher than the network data volume during gradient aggregation across all training rounds, i.e., $B\cdot(\sum_{\hopiter=1}^{\num-1}H_{\hopiter} + H_0) \geq M\cdot R$. This is true for the real-world datasets we evaluated. In this case, $\gamma$ increases continuously with $B$ until it approximately stabilizes at $\Theta(R)$. In our extended analysis 
 (see table \ref{tab:gamma_comp_products} and \ref{tab:gamma_comp_reddit} in Appendix \ref{appdx:commanalysis}), we will show that the measured data volume while training on our distributed testbed matches with our analysis for $\gamma$.

In conclusion, we have shown that Lazy message passing training saves up to $\Theta(R)\times$ network data volume compared to the standard training. Under mild conditions that are true for real-world datasets, we show that the gap (ratio) in network data volume between the two training procedures increases with the number of boundary nodes per worker until a maximum of $\Theta(R)$ for a large number of boundary nodes.  

\noindent{\bfseries Note.} $\gamma$ can be computed even before training the GNN to assess the savings in communication costs prior to training.

\begin{figure}
    \centering
    \includegraphics[scale=0.30]{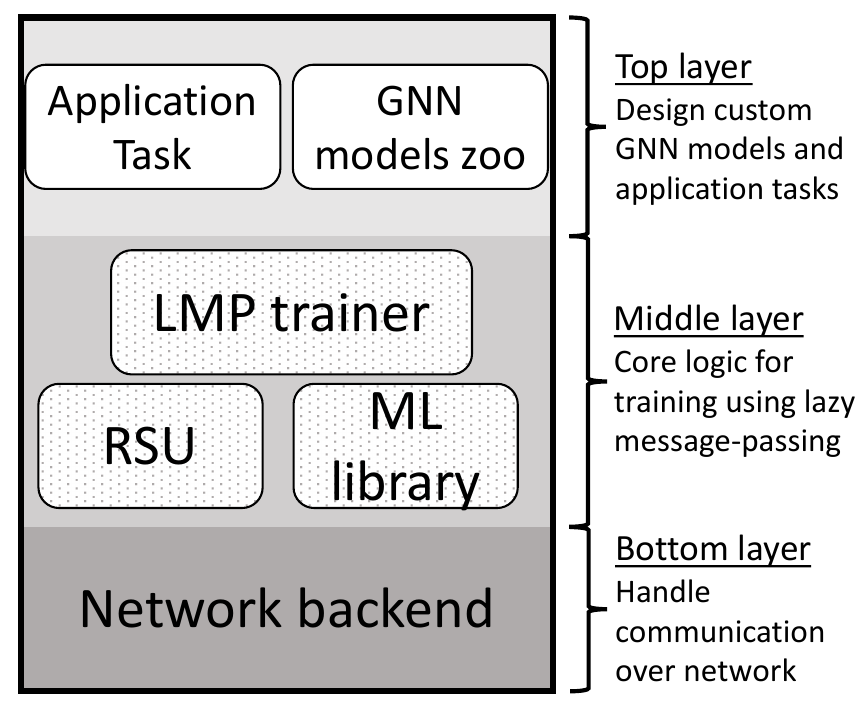}
    \caption{\tool's layers of abstraction.}
    \label{fig:retexoabs}
\end{figure}

\subsection{\tool}
\label{sec:retexo-arch}

\tool is a framework to train GNNs end-to-end using our lazy message passing strategy. We describe the framework conceptually using $3$ high-level abstract layers (see Figure~\ref{fig:retexoabs}). 


\paragraph{Bottom layer.} This has the Backend communications component that handles the communication over TCP/IP, bluetooth, or similar networking backends. Mainly, it provides three API calls, \texttt{init}, \texttt{send}, \texttt{recv}. The \texttt{init} can be used to initialize the communication channels between workers. Subsequently, \texttt{send} and \texttt{recv} can be used to deliver messages over these communication channels.

\paragraph{Middle layer.} There are three components in this layer. The Reduce-sync-update (RSU) is responsible for three functionalities, aggregating (reducing) the gradients, synchronizing and scheduling the training process, and sharing the updated intermediate embeddings of the boundary nodes. The ML library is used to perform training and inference on the available hardware (CPU/GPU). Finally, LMP trainer is built using the ML library and the RSU components. It implements the logic of lazy message passing.

\paragraph{Top layer.} The GNN model zoo captures the GNN models that are built using the API provided by the  ML library such as PyTorch or other third-party libraries such as Deep Graph Library (DGL) which is itself built on top of PyTorch. Custom GNN models can be written by the developer too.
The Application Task component handles the tasks that are necessary for the GNN application such as creating data splits to train for node classification, defining the loss function, evaluation metrics, and the hyperparameters.  Finally, the Application Task component can also be used to define custom aggregating functions to aggregate the gradients at the master. By default, the aggregation function is a simple average of gradients. However, one could also implement custom aggregation functions such as defenses against data poisoning, and training for better robustness~\cite{liu2022federated}

\subsection{Implementation}
\label{sec:implementation}
We implement the current version of \tool on top of three libraries, PyTorch, PyTorch Distributed, and DGL~\cite{wang2019dgl}. PyTorch Distributed provides our Backend communications component, PyTorch is the ML library, and DGL is used in the top layer to support functionalities such as splitting the graph data and building the GNN models. 
The RSU component consists of three API \texttt{MultiThreadedReducer}, \texttt{sync\_model}, and \texttt{sync\_embeddings}. The \texttt{MultiThreadedReducer} API handles the sending of the gradients to the master, aggregating them, and sending them back to the workers. Workers use the \texttt{sync\_embeddings} API to send and receive the trained embeddings of the boundary nodes with their peers. Similarly, the \texttt{sync\_model} is used to sync the model with the best validation accuracy among all the workers. These APIs are used by the LMP trainer to conduct training while ensuring that all the workers have a consistent state. \tool can be modified to fit any application setup for graph learning by configuring the bottom and top layer components. For instance, we show that \tool can be deployed even in highly-decentralized mobile setups by building a custom Bluetooth-based Network backend on top of the Flower federated learning library. We discuss this further in Section~\ref{sec:evaldecentralized}. 

\paragraph{Measuring network data cost.} We instrument \tool's codebase to log the amount of data sent and received over all communication channels. To confirm that our instrumentation is correct, we use Wireshark to monitor the network and save the network packets sent and received on all communication channels. The difference between the measurements obtained from instrumentation and external monitoring is negligible (smaller than $0.1\%$) and it exists due to the additional network headers that the instrumentation ignores. Furthermore, our instrumentation allows us to measure network data volume for every worker irrespective of the underlying architecture on which the workers have been deployed. We deploy multiple workers per GPU and multiple workers on a single machine for instance. Workers within a single machine  communicate over PCIe and across machines over Ethernet in an internal network. Nevertheless, our data volume measurements will still represent how much each worker would communicate with the others in any edge-partitioned distributed setup where all workers may communicate over the Internet or other bandwidth constrained channels.

\section{Evaluation}
\label{sec:eval_new}

We have three main evaluation objectives.
\begin{enumerate}
    \item How well do \toolgnns perform compared to GNNs on supervised node classification tasks?
    \item What are the network communication characteristics of training GNNs with \tool and how do they compare with the baseline training strategy of GNNs?
    \item How \tool's training strategy compares with state-of-the-art optimization boundary node sampling (BNS)?
\end{enumerate}


\subsection{Evaluation Setup}
\label{sec:eval_setup}

\begin{table*}[t]
\caption{Micro-F1 scores for node classification in the transductive setting.}
\label{tab:transresults}
\resizebox{\textwidth}{!}{%
\begin{tabular}{l|l|l|l|l|l|l|l}
Method     & \cora            & \seer        & \pubmed          & \fbpage  & \lastfm & \reddit & \products \\ \hline
\gcn   &     $0.820 \pm 0.005$ & $0.725 \pm 0.006$ & $0.838 \pm 0.001$ & $0.917 \pm 0.001$ & $0.846 \pm 0.002$ & $0.925 \pm 0.000$ & $0.713 \pm 0.000$ \\
\toolgcn& $0.828 \pm 0.006$ & $0.739 \pm 0.009$ & $0.850 \pm 0.002$ & $0.916 \pm 0.003$ &  $0.843 \pm 0.005$ & $0.939 \pm 0.000$ &  $0.791 \pm 0.001$\\
\sage  & $0.826 \pm 0.004$ & $0.743 \pm 0.010$ & $0.867 \pm 0.002$ & $0.925 \pm 0.001$ & $0.835 \pm 0.004$ & $0.952 \pm 0.000$ & $0.781 \pm 0.000$ \\
\toolsage & $0.825 \pm 0.004$ & $0.766 \pm 0.008$ & $0.861 \pm 0.002$ & $0.910 \pm 0.002$ &  $0.825 \pm 0.007$ & $0.957 \pm 0.004$ & $0.792 \pm 0.001$\\
\gat  & $0.807 \pm 0.007$ & $0.714 \pm 0.011$ & $0.833 \pm 0.002$ & $0.912 \pm 0.003$ & $0.842 \pm 0.004$ & $0.925 \pm 0.000$ & $0.707 \pm 0.000$\\
\toolgat  & $0.818 \pm 0.006$ & $0.738 \pm 0.010$ & $0.830 \pm 0.002$ & $0.913 \pm 0.003$ & $0.832 \pm 0.005$ & $0.944 \pm 0.003$ &  $0.748 \pm 0.001$
\end{tabular}
}
\end{table*}


\paragraph{Datasets \& Splits.}
We consider $7$ widely-used GNN benchmarks of varying nature and sizes ($\approx$ $9K-100M$ edges) to measure \tool's performance. They include citation networks \cora, \seer, and \pubmed; social networks, \lastfm, \fbpage, and \reddit; and an Amazon product co-purchasing network called \products. We perform node classification in the commonly studied transductive setting~\cite{hamilton2017inductive} for all datasets where the graph is fixed and not changing between training, validation, and testing phases. Due to space constraints, we provide additional details regarding the datasets and their splits in Appendix~\ref{appdx:evalsetup}.

\paragraph{Models Architectures \& Baselines.} We train three most popular GNNs used in practice \gcn~\cite{kipf2016semi}, \sage~\cite{hamilton2017inductive}, and \gat~\cite{velivckovic2017graph}. By training these GNNs using \tool we obtain corresponding \toolgnns, i.e., \toolgcn, \toolsage, and \toolgat. First, we compare \tool with the baseline training strategy. We refer to the GNNs trained using the baseline strategy simply by the GNN names.

On the two large datasets \reddit and \products, we further compare our optimization strategy with boundary node sampling (BNS) since it is the state-of-the-art solution that applies to our problem setup to the best of our knowledge~\cite{wan2022bns}. We choose the sampling rate for sampling the boundary nodes to be $0.1$ as is suggested in their paper. We will also show why lower sampling rates than that might not be desirable. Thus the second set of baselines are \gcna, \sagea, and \gata. Finally, we choose the hyperparameters to train GNNs based on the ones mentioned in the prior work~\cite{wan2022bns,hamilton2017inductive} (see Appendix~\ref{appdx:evalsetup} for more details).

\paragraph{System Specifications.} We conduct experiments on two machines each with $4$ A40 GPUs (48 GB) and AMD EPYC 7443P $24$-Core Processors@$2.8$GHz with $252$ GB memory. All of them are connected via PCIe4x16 on a single machine. The two machines are interconnected via Ethernet on an internal network with $10$ Gbps bandwidth that is fully available to these machines.

\paragraph{Evaluation Outline.}
\tool embodies a novel training procedure, hence, it is important to compare the node classification performance of \toolgnns and GNNs trained with standard training. Therefore, we first report on the classification accuracy on all datasets considered. On all small datasets, we train \toolgnns and GNNs for $100$ rounds (epochs) and report the test accuracy achieved at that training round where the best validation accuracy is achieved. For large datasets, we train the GNNs first and observe when they start to converge to give their respective best accuracies. We find that this happens on both \reddit and \products at around $200$ training rounds. Hence, we fix the number of rounds to $200$ and train their corresponding \toolgnns, i.e., we train each layer of a $\num$-layer GNN for $200$ rounds thus effectively performing $200\cdot\num$ training rounds. 

We observe that the time taken to train a $\num$-layer GNN with standard training in $R$ rounds is comparable to or higher than that of training all layers in a GNN using \tool for $R$ rounds.
We report all accuracies over $10$ runs.
\begin{figure}[t]
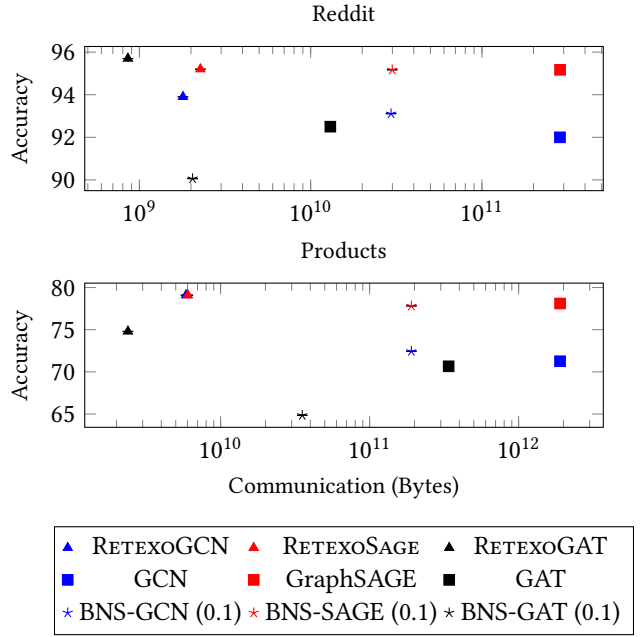

    \begin{tikzpicture}
        \begin{groupplot}[group style = {group size = 1 by 2, horizontal sep = 20pt, vertical sep = 35pt, xlabels at=edge bottom, ylabels at=edge bottom},
                width=\columnwidth,
                height = 3.5cm,
                x label style={font=\small},
                y label style={font=\small},]
            \input{figures/reddit_acc_vs_dv_2}
            \input{figures/ogbn-products_acc_vs_dv_2}
        \end{groupplot}
        \node at ($(group c1r2) + (0.0cm,-3.0cm)$) {\ref{grouplegend1}};
    \end{tikzpicture}
    \caption{Comparison of accuracy vs total data volume transferred  per client for $2$ workers (log scale).}
    \label{fig:acc_vs_comm_fed}
\end{figure}
For the network data cost experiments we choose the two larger datasets \reddit ($\approx$ $233K$ nodes and $114M$ edges) and \products ($\approx$ $2.7M$ nodes and $62M$ edges). We perform distributed training for \toolgnns and the two baselines across various decentralization levels. We first compare the network data volume at low decentralization levels, i.e., $\#$ of workers $(N) = 2, 4$, and report the data volume sent over the network. We repeat these experiments for higher decentralization levels of $N=6, 8, 16$, and $24$ for \reddit whereas $6, 8$ for \products. We do not choose more workers for \products due to limited computational resources available on our test machines. However, we expect conclusions drawn from our results will extend to larger workers and to larger graph sizes, for instance, beyond $100M$ nodes.

\subsection{Accuracy}
\label{sec:eval_accuracy}
We report the best accuracies achieved by \toolgnns and their corresponding GNNs in Table~\ref{tab:transresults}. The accuracy of \toolgnns are $\pm 1\%$  of their corresponding standard GNNs, and in a few configurations, \toolgnns performs better by up to $4\%$ and $8\%$. Specifically, on the \products dataset, \toolgcn and \toolsage have accuracy close to $79\%$ that are higher than the best reported accuracies for their GNN counterparts on the public leaderboard\footnote{Products Learderboard: \url{https://ogb.stanford.edu/docs/leader_nodeprop/\#ogbn-products}---\gcn ($75.6\%$) and \sage ($78.2\%$)}. This shows that training GNNs with lazy message passing has as good or better classification performance as standard training.

\begin{figure}[t]
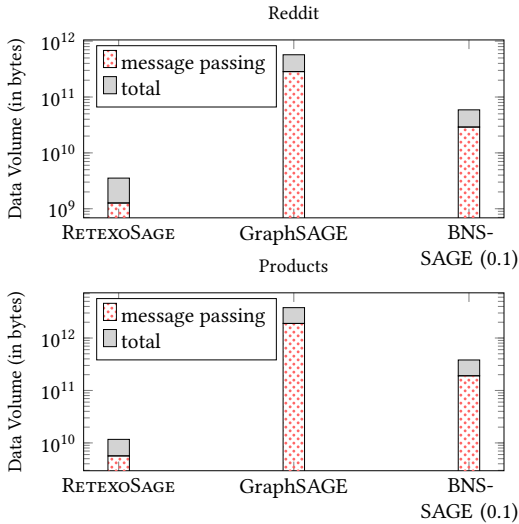

    \resizebox{0.85\columnwidth}{!}{%
    \begin{tikzpicture}
        \begin{groupplot}[group style = {group size = 1 by 2, horizontal sep = 100pt, vertical sep = 35pt, xlabels at=edge bottom, ylabels at=edge bottom},
                width=1/2.1*\textwidth,
                height = 4.5cm,
                x label style={font=\small},
                y label style={font=\small},
                ]
            \input{figures/reddit_finer_comm_graphsage_2}
            \input{figures/products_finer_comm_graphsage_2}
        \end{groupplot}
    \end{tikzpicture}
    }
    \caption{Total data volume during message passing as compared to the total data volume (log scale).}
    \label{fig:finer_comm_graphsage_2_fed}
\end{figure}
\subsection{Data Volume for low Decentralization}
\label{sec:commeffl}

To measure the network data costs we start with the lowest levels of decentralization with only $2$ workers. This setup is supposed to be the least expensive setup for communication for any architecture since the majority of the graph's edges would be within the workers itself and gradients across only two workers need to be synced. Thus, reducing the total data volume communicated during message passing. 

We plot accuracy vs. total data volume sent over the network per worker during the entire training process in Figures~\ref{fig:acc_vs_comm_fed} for all GNN architectures. We consider only $2$ workers for all configurations except for \gat-based models in Products due to limited GPU memory.
We find that standard training of GNNs requires sending up to $285$ GB for \reddit and $1.9$ TB for \products over the network. In comparison, training GNNs with \tool only requires up to $2.2$ GB and $6$ GB to be sent over the network.

The state-of-the-art system, BNS-GNNs ($0.1$), requires sending up to $29.5$ GB and $191$ GB over the network per worker. Therefore, training \toolgnns requires up to $315\times$ and $32\times$ lower data volume than all baselines on \products dataset. At the same time, \toolgnns perform as good as (or better) than the compared baselines.

To understand the stark difference in data volume required for \toolgnns vs other baselines, we plot the data volume required for gradient aggregation and message passing separately in Figure~\ref{fig:finer_comm_graphsage_2_fed}. The data volume required for message passing is the dominating factor in this setup. For instance, more than $99.5\%$ of the data is sent during the message passing phase for standard GNNs and more than $96\%$ for BNS-GNNs ($0.1$). It is as expected, since in every training round, only one gradient per worker is sent over the network during the gradient aggregation phase whereas during message passing features and gradients corresponding to all boundary nodes are exchanged between workers. In contrast, in \tool there are only $\num$ message passing rounds for a $\num$-layer GNN, therefore, the data volume required during both phases are comparable to each other.
\vspace{0.1cm}
\begin{mdframed}[backgroundcolor=mygray]
   The network data exchanged to train \toolgnn is $1-2$ orders of magnitude less than with standard training and with BNS-GCN (0.1) for real-world graphs.
\end{mdframed}

\input{figures/decentralization_communication_reddit}
\subsection{Data Volume for High Decentralization}
\label{sec:commeffh}

We increase the number of workers from $2$ to $24$ for \reddit and $2$ to $8$ for \products to measure the total data volume and we report the values in Figure~\ref{fig:total_dv} for \sage-based architectures. On each bar, we present a number which represents the ratio of total data volume of that bar to the corresponding bar of \toolgnn. \gcn and \gat-based architectures  have similar plots, hence elided here for brevity. We see that as the number of workers increases, the advantage offered by \toolgnns over other baselines increases. For instance, the gap between training \toolsage and \sage increases from $315\times$ to $321\times$ for the \products dataset.

This is expected as per our analysis for $\gamma$ (see Section~\ref{sec:commanalysismain}) which is the ratio of network data volume for baseline strategy and the lazy message passing strategy. Specifically, in all of our evaluated configurations, as decentralization increases more nodes per worker become boundary nodes for other workers. Therefore, the total data volume sent over the network during each message passing round also increases and still dominates the gradient aggregation phase.
\vspace{0.1cm}
\begin{mdframed}[backgroundcolor=mygray]
   The gap in data volume between training GNNs using \tool vs baselines increases with decentralization.
\end{mdframed}

\begin{table}[t]
    \centering
    \caption{Accuracy with increasing sampling.}
    \begin{tabular}{c|c|c}
       Architecture & \reddit  & \products \\\hline
        \toolgat & 0.942 & 0.749 \\
        \gat & 0.925 & 0.704 \\
        \gata & 0.903  & 0.648 \\
        \gatb & 0.404 & 0.348 \\\hline
    \end{tabular}
    \label{tab:gataccsampling}
\end{table}

\subsection{Accuracy vs Sampling}
\label{sec:evalacc}

\tool offers an orthogonal opportunity to optimize network costs compared with prior approaches, prominently sampling as in BNS. It is nonetheless worth comparing how the two differ.
Until now we have compared BNS at $0.1$ sampling rate, as also considered in prior work. We can further measure what happens if we sub-sample the boundary nodes even more aggressively. We observe that if the sampling rate is $0.01$ instead of $0.1$, the data volume during message passing reduces by $10\times$ for BNS as expected, making it within the same order of magnitude as data efficient as \tool. Figure~\ref{fig:total_dv} shows this. But, crucially, such aggressive sampling results in a significant loss of GNN accuracy. We illustrate this for \gat-based architectures, since they are the most affected, in Table~\ref{tab:gataccsampling}. All architectures are trained for the same number of rounds where the original \gat converges. For both datasets, the accuracy drops sharply by up to $50\%$ for a sampling rate of $0.01$ and drops significantly ($5.6\%$) even for a sampling rate of $0.1$ for \products.
In contrast, \tool's training procedure does not require to sample boundary nodes. So its accuracy does not change with any level of decentralization. 



\begin{table*}[ht]
    \centering
    \caption{Time taken (in seconds) for total training and message passing (MP) with varying latency}
    \resizebox{0.75\textwidth}{!}{
    \begin{tabular}{|c|c|c|c|c|c|c|c|c|}
        \hline
         & \multicolumn{2}{c|}{Latency (0 ms)} & \multicolumn{2}{c|}{Latency (50 ms)} & \multicolumn{2}{c|}{Latency (100 ms)} & \multicolumn{2}{c|}{Latency (200 ms)} \\
        \cline{1-9}
         {\bfseries Reddit} & Total & MP & Total & MP & Total & MP & Total & MP\\
        \hline
        \toolsage & 41.3 & 1.7 & 103.7 & 1.5 & 181.9 & 1.5 & 339.8 & 1.5 \\
        \hline
        \sage & 290.6 & 262.0 & 325.2 & 274.8 & 409.4 & 339.5 & 802.1 & 665.7 \\
        \hline
        \sagea & 43.9 & 32.6 & 113.4 & 82.8 & 193.3 & 142.9 & 370.5 & 279.8 \\
        \hline
        \textbf{Products} &  &  &  &  &  &  &  & \\
        \hline
        \toolsage & 76.0 & 6.3 & 112.2 & 6.1 & 158.0 & 6.0 & 275.3 & 6.0 \\
        \hline
        \sage & 1681.2 & 1528.8 & 1692.8 & 1526.4 & 1711.6 & 1519.3 & 2176.4 & 1912.4 \\
        \hline
        \sagea & 205.3 & 168.9 & 233.9 & 180.5 & 289.1 & 214.6 & 407.7 & 293.9 \\
        \hline
    \end{tabular}
    }
    \label{tab:latency_comparison}
\end{table*}
\subsection{Training Time}
\label{sec:eval_train_time}

The secondary impact of reducing the network data volume is on the training time. Note that our testbed is similar to the prior works~\cite{wan2022bns,gandhi2021p3} and our baselines have been optimized to minimize the training time. On our testbed, the total time taken to train a GNN is split between the local computations on GPUs and network communication with the latter being the bottleneck. We assume a centralized data center setup and do not throttle the bandwidth to give full advantage to our baselines. We also add different network latencies, from $0$ to $200$ ms, to simulate workers in different geographical regions connected over dedicated high-bandwidth communication channels. The latencies are in line with those reported for geographically separated Azure servers as reference.~\footnote{\url{https://learn.microsoft.com/en-us/azure/networking/azure-network-latency?tabs=Americas\%2CWestUS}} Table~\ref{tab:latency_comparison} presents the total training time to train \sage  on $4$ workers for $200$ training rounds.

\tool trains \sage faster than both the baselines in all evaluated configurations for \reddit and \products datasets. At no additional latency simulated, \tool trains \sage by up to $7\times$ and $22\times$ faster than the baseline procedure on \reddit and \products datasets respectively. Further, compared to BNS at a sampling rate $0.1$, \tool is slightly faster on \reddit and $2.7\times$ faster on \products. \tool spends almost no time during the message passing rounds as expected. Most of its training time is spent during the local training phases and in gradient aggregation. In contrast, the baselines spend most of their time during the message passing rounds.

As latency increases, the training time for all procedures increases. At an  additional latency of $50 ms$, \tool is up to $15\times$ faster than the baseline procedure and $2\times$ faster than BNS. With increasing latencies, while \tool maintains the advantage, though the improvement in training time is not linear compared to other baselines.  This is primarily because the network data volume is much smaller during the gradient aggregation phase than during the message passing rounds for all training procedures. 
%
Therefore, message-passing rounds are bandwidth-bound as opposed to latency-bound, whereas gradient aggregation is latency-bound.~\footnote{at least for network throughput higher than $32$ Mbps per worker for GNNs with less than $10^6$ parameters ($4$ MB).}
\tool has much fewer message passing rounds compared to the baselines, but  $\num\cdot\numrounds$ gradient aggregation rounds compared to  $\numrounds$ rounds in the baselines.
So, the impact of increasing latency is more pronounced on \tool, since more of its internal phases are latency-bound.
In contrast, the baselines have much more bandwidth-bound message-passing phases, and hence, increasing latency has a less pronounced end effect on their resulting training time.
Nevertheless, even at $200$ ms latency, \tool is up to $8\times$ faster than the standard training and $1.5\times$ faster than BNS.
\vspace{0.2cm}
\begin{mdframed}[backgroundcolor=mygray]
   \tool trains GNNs faster than both baselines across all evaluated configurations. Specifically, it is up to $22\times$ and $2.7\times$ faster than the baseline procedure and BNS ($0.1$) respectively at low network latencies. 
\end{mdframed}

\subsection{Extending to High Decentralization}
\label{sec:evaldecentralized}

All of the aforementioned results for \tool indicate that it scales gracefully with increasing decentralization due to a small number of message passing rounds. Now, we also demonstrate the same by training GNNs using \tool in a fully-distributed setup. In the fully-distributed setup, workers can be mobile or stationary IoT devices. Mainly, these devices are bandwidth constrained and may be in mobile environments with unreliable networks. We consider a specific application scenario of proximity networks. In such networks, workers share information over low-bandwidth bluetooth channels or using WiFi-Direct with each other when in close proximity. Such networks are widely used for contact-tracing, and sharing content with contacts (on Apple Airdrop and Android's Nearby Share). To simulate these applications, we use two Raspberry Pis (RPi)s as mobile devices and train a \gcn using \tool on them. The RPis communicate with each other over Bluetooth and communicate with a server over Wi-Fi (Internet). The server coordinates the training of \gcn on these mobile workers.

We split the \cora dataset equally among them to train a $2$-layer \gcn on these devices. We provide additional details on the setup and the training procedure in Appendix~\ref{appdx:feasibility}. We build \tool for fully-distributed setup  on top of a popular federated learning framework called Flower~\cite{beutel2020flower} (see Appendix~\ref{appdx:feasibility} for more details). It offers easier tools to build for highly decentralized setups\footnote{\url{https://github.com/adap/flower}}. The framework provides worker-to-master channels and we implement our own worker-to-worker communication channels over Bluetooth. Quantitatively, we report the performance characteristics we observe during training.

The total data communicated for gradient aggregation, including the headers added by the network protocols, to train the first layer is $585$ MB. In total, $0.08$ MB is communicated over Bluetooth for the message passing rounds. The maximum RAM consumed on each RPi is $165.44$ MB. \tool requires $63.28$ seconds for message passing, which includes the connection time, requesting specific nodes' embeddings, and exchanging the requested ones. Overall, the RAM consumed is well within the constraints of typical embedded systems ($<1$ GB RAM). 
%
Finally, the fully-distributed setup may have only one inner graph node per worker, for instance, when the worker is a mobile phone of a user (node) in a social or contact network. Resource-wise, training \toolgnns is even more practical in such setups, unlike above, since each RPi will consume much less RAM and communicate much less data. For instance, two RPis require only $0.6$ seconds to connect and exchange a single node's embedding. This conclusion is independent of graph size since the resources needed per RPi do not scale with the number of nodes.
\vspace{0.2cm}
\begin{mdframed}[backgroundcolor=mygray]
   \tool is the first framework to demonstrate training GNNs in fully-decentralized setups corresponding to mobile and edge networks.
\end{mdframed}
\section{Related Work}
\label{sec:related}

With the rising prominence of GNNs and their applications on industry-scale large graphs, recently, many works have addressed the problem of efficiently training them  by distributing the graph data and GNN computation.

\paragraph{Efficient GNN Training.}
Many works focus on reducing the end-to-end training time in a centralized datacenter-like setup. They can be further split in to  systems designed for single-machine~\cite{hu2020featgraph,ma2019neugraph,lin2020pagraph,jangda2021accelerating} and multi-machine settings~\cite{zheng2020distdgl,cai2021dgcl,gandhi2021p3,wan2022bns,zhu2019aligraph,jia2020improving,md2021distgnn,thorpe21dorylus,zheng2022bytegnn,liu2023bgl}.

Single-machine systems assume that the graph data is located on a single machine, possibly with multiple CPUs/GPUs to train the GNNs. FeatGraph~\cite{hu2020featgraph} generates optimized kernels for both CPUs and GPUs that execute GNN-based operations such as message passing faster. NeuGraph~\cite{ma2019neugraph} optimizes CPU-GPU data transfer to execute GNN computation graphs, by proposing a streaming scheduler that maximizes the overlap between computation and data transfer. Other works optimize the speed of minibatch sampling for graphs on GPUs~\cite{lin2020pagraph,jangda2021accelerating}. Single-machine systems do not consider network data cost as a bottleneck. Their optimizations are orthogonal and complementary to our work, i.e., they can be implemented for \tool to improve its performance during local training phase on individual machines.

Multi-machine systems assume that the graph data can be partitioned and stored on multiple networked machines. They all aim to train GNNs fast and propose several optimizations to do so. Unlike the single-machine systems, they raise the issue of high network data costs incurred during message passing rounds being the main bottleneck. Roc~\cite{jia2020improving}, DistGNN~\cite{md2021distgnn}, and DGCL~\cite{cai2021dgcl} follow the standard training strategy but propose various competing graph partition strategies to reduce the network data costs. AliGraph~\cite{zhu2019aligraph}, P3~\cite{gandhi2021p3}, DistDGL~\cite{zheng2020distdgl}, BGL~\cite{liu2023bgl}, and ByteGNN~\cite{zheng2022bytegnn} combine graph partitioning and mini-batch sampling in different ways to reduce the network data cost. At their core, they do not follow the standard training procedure that we have described. In every training round, the workers subsample a set of nodes (mini-batch) and then fetch the entire K-hop neighborhoods of those nodes from other workers to train the GNN. This essentially replicates raw features and edges of inner nodes, that are beyond the boundary nodes, of other workers. It is evident that all of these systems have been designed for the centralized setup and none of the aforementioned optimizations would apply in the distributed setups with data decentralization constraints. Further, their definition of ``efficiency'' corresponds to training speed only since network data volume in the centralized setup is not costly (financially). In contrast, we primarily focus on reducing the network data volume since it is the root cause for expensive training in terms of money and time. Finally, to the best of our knowledge, BNS-GCN is the only system to perform the standard training procedure with optimization that reduces network data costs in our setup (see Section~\ref{sec:prioropt}).

\paragraph{Alternative Distributed Setups.}
Prior works have trained GNNs in other distributed setups with different ways of storing graph data, and with different notions of privacy from our setup. These works~\cite{chen2021fedgl,wang2020graphfl,zheng2021asfgnn,zhang2021subgraph,he2021fedgraphnn,wang2022federatedscope} propose distributed setups where cross-worker communication is not performed at all to train GNNs. They assume that the entire graph is replicated across workers~\cite{wang2020graphfl} or ignore the cross-worker edges altogether, thus, effectively conducting training similar to BNS-GCN with $0$ sampling rate~\cite{he2021fedgraphnn,wang2022federatedscope}. Further, ~\cite{chen2021fedgl} and~\cite{zhang2021subgraph} make the master help the workers to predict missing graph information from other workers apart from just aggregating gradients. These works are motivated differently, such as dealing with non-IID graph data partitions, and do not focus on training GNNs efficiently on large graphs. Their techniques do not extend to the edge-partitioned distributed setup, and also to high decentralization levels.

Finally, GNNs have been trained with differential privacy guarantees in distributed setups which allows workers to send node features and edges to the master with adequate anonymization~\cite{sajadmanesh2021locally, kolluri2022lpgnet,zhu2023blink}. However, these works assume that partial graph data, edges or features, is accessible to the master worker and privacy is preserved for the inaccessible data. Further, differential privacy can be detrimental to classification accuracy---the trained GNNs in the aforementioned works have performance loss of ($>10\%$) compared to non-private training. We believe that \tool provides a baseline platform to implement and test differentially private training of GNNs and leave it for the future work.


\section{Conclusion}
\label{sec:conclusion}

We have presented a new framework \tool to train GNNs on distributed machines, for all decentralization levels, while eliminating the severe communication bottlenecks. Unlike prior works, \tool does not need the machines to share raw graph edges or node features with any centralized coordinator or other machines. \tool reduces the network data costs by up to $2$ orders of magnitude compared to state-of-the-art baselines to train GNNs on benchmarked datasets. To achieve this, \tool implements a novel training strategy, called lazy message passing, where the key idea is to reduce the number of message passing rounds required to train the GNN compared to the standard training procedure.


\bibliography{paper}


\begin{thebibliography}{72}


\ifx \showCODEN    \undefined \def \showCODEN     #1{\unskip}     \fi
\ifx \showDOI      \undefined \def \showDOI       #1{#1}\fi
\ifx \showISBNx    \undefined \def \showISBNx     #1{\unskip}     \fi
\ifx \showISBNxiii \undefined \def \showISBNxiii  #1{\unskip}     \fi
\ifx \showISSN     \undefined \def \showISSN      #1{\unskip}     \fi
\ifx \showLCCN     \undefined \def \showLCCN      #1{\unskip}     \fi
\ifx \shownote     \undefined \def \shownote      #1{#1}          \fi
\ifx \showarticletitle \undefined \def \showarticletitle #1{#1}   \fi
\ifx \showURL      \undefined \def \showURL       {\relax}        \fi
\providecommand\bibfield[2]{#2}
\providecommand\bibinfo[2]{#2}
\providecommand\natexlab[1]{#1}
\providecommand\showeprint[2][]{arXiv:#2}

\bibitem[\protect\citeauthoryear{Beutel, Topal, Mathur, Qiu, Fernandez-Marques, Gao, Sani, Kwing, Parcollet, Gusmão, and Lane}{Beutel et~al\mbox{.}}{2020}]%
        {beutel2020flower}
\bibfield{author}{\bibinfo{person}{Daniel~J Beutel}, \bibinfo{person}{Taner Topal}, \bibinfo{person}{Akhil Mathur}, \bibinfo{person}{Xinchi Qiu}, \bibinfo{person}{Javier Fernandez-Marques}, \bibinfo{person}{Yan Gao}, \bibinfo{person}{Lorenzo Sani}, \bibinfo{person}{Hei~Li Kwing}, \bibinfo{person}{Titouan Parcollet}, \bibinfo{person}{Pedro PB~de Gusmão}, {and} \bibinfo{person}{Nicholas~D Lane}.} \bibinfo{year}{2020}\natexlab{}.
\newblock \showarticletitle{Flower: A Friendly Federated Learning Research Framework}.
\newblock \bibinfo{journal}{{\em arXiv preprint arXiv:2007.14390\/}} (\bibinfo{year}{2020}).
\newblock


\bibitem[\protect\citeauthoryear{Bhagoji, Chakraborty, Mittal, and Calo}{Bhagoji et~al\mbox{.}}{2019}]%
        {bhagoji2019analyzing}
\bibfield{author}{\bibinfo{person}{Arjun~Nitin Bhagoji}, \bibinfo{person}{Supriyo Chakraborty}, \bibinfo{person}{Prateek Mittal}, {and} \bibinfo{person}{Seraphin Calo}.} \bibinfo{year}{2019}\natexlab{}.
\newblock \showarticletitle{Analyzing federated learning through an adversarial lens}. In \bibinfo{booktitle}{{\em ICML}}.
\newblock


\bibitem[\protect\citeauthoryear{Bojchevski and G{\"u}nnemann}{Bojchevski and G{\"u}nnemann}{2019a}]%
        {bojchevski2019adversarial}
\bibfield{author}{\bibinfo{person}{Aleksandar Bojchevski} {and} \bibinfo{person}{Stephan G{\"u}nnemann}.} \bibinfo{year}{2019}\natexlab{a}.
\newblock \showarticletitle{Adversarial attacks on node embeddings via graph poisoning}. In \bibinfo{booktitle}{{\em ICML}}.
\newblock


\bibitem[\protect\citeauthoryear{Bojchevski and G{\"u}nnemann}{Bojchevski and G{\"u}nnemann}{2019b}]%
        {bojchevski2019certifiable}
\bibfield{author}{\bibinfo{person}{Aleksandar Bojchevski} {and} \bibinfo{person}{Stephan G{\"u}nnemann}.} \bibinfo{year}{2019}\natexlab{b}.
\newblock \showarticletitle{Certifiable robustness to graph perturbations}.
\newblock \bibinfo{journal}{{\em NeurIPS\/}} (\bibinfo{year}{2019}).
\newblock


\bibitem[\protect\citeauthoryear{Bonawitz, Eichner, Grieskamp, Huba, Ingerman, Ivanov, Kiddon, Kone{\v{c}}n{\`y}, Mazzocchi, McMahan, et~al\mbox{.}}{Bonawitz et~al\mbox{.}}{2019}]%
        {bonawitz2019towards}
\bibfield{author}{\bibinfo{person}{Keith Bonawitz}, \bibinfo{person}{Hubert Eichner}, \bibinfo{person}{Wolfgang Grieskamp}, \bibinfo{person}{Dzmitry Huba}, \bibinfo{person}{Alex Ingerman}, \bibinfo{person}{Vladimir Ivanov}, \bibinfo{person}{Chloe Kiddon}, \bibinfo{person}{Jakub Kone{\v{c}}n{\`y}}, \bibinfo{person}{Stefano Mazzocchi}, \bibinfo{person}{Brendan McMahan}, {et~al\mbox{.}}} \bibinfo{year}{2019}\natexlab{}.
\newblock \showarticletitle{Towards federated learning at scale: System design}.
\newblock \bibinfo{journal}{{\em MLSys\/}} (\bibinfo{year}{2019}).
\newblock


\bibitem[\protect\citeauthoryear{Bonawitz, Ivanov, Kreuter, Marcedone, McMahan, Patel, Ramage, Segal, and Seth}{Bonawitz et~al\mbox{.}}{2017}]%
        {bonawitz2017practical}
\bibfield{author}{\bibinfo{person}{Keith Bonawitz}, \bibinfo{person}{Vladimir Ivanov}, \bibinfo{person}{Ben Kreuter}, \bibinfo{person}{Antonio Marcedone}, \bibinfo{person}{H~Brendan McMahan}, \bibinfo{person}{Sarvar Patel}, \bibinfo{person}{Daniel Ramage}, \bibinfo{person}{Aaron Segal}, {and} \bibinfo{person}{Karn Seth}.} \bibinfo{year}{2017}\natexlab{}.
\newblock \showarticletitle{Practical secure aggregation for privacy-preserving machine learning}. In \bibinfo{booktitle}{{\em proceedings of the 2017 ACM SIGSAC Conference on Computer and Communications Security}}.
\newblock


\bibitem[\protect\citeauthoryear{Bottou, Curtis, and Nocedal}{Bottou et~al\mbox{.}}{2018}]%
        {bottou2018optimization}
\bibfield{author}{\bibinfo{person}{L{\'e}on Bottou}, \bibinfo{person}{Frank~E Curtis}, {and} \bibinfo{person}{Jorge Nocedal}.} \bibinfo{year}{2018}\natexlab{}.
\newblock \showarticletitle{Optimization methods for large-scale machine learning}.
\newblock \bibinfo{journal}{{\em SIAM review\/}} (\bibinfo{year}{2018}).
\newblock


\bibitem[\protect\citeauthoryear{Cai, Yan, Wu, Ma, Cheng, and Yu}{Cai et~al\mbox{.}}{2021}]%
        {cai2021dgcl}
\bibfield{author}{\bibinfo{person}{Zhenkun Cai}, \bibinfo{person}{Xiao Yan}, \bibinfo{person}{Yidi Wu}, \bibinfo{person}{Kaihao Ma}, \bibinfo{person}{James Cheng}, {and} \bibinfo{person}{Fan Yu}.} \bibinfo{year}{2021}\natexlab{}.
\newblock \showarticletitle{DGCL: an efficient communication library for distributed GNN training}. In \bibinfo{booktitle}{{\em EuroSys}}.
\newblock


\bibitem[\protect\citeauthoryear{Chen, Hu, Xu, and Zheng}{Chen et~al\mbox{.}}{2021}]%
        {chen2021fedgl}
\bibfield{author}{\bibinfo{person}{Chuan Chen}, \bibinfo{person}{Weibo Hu}, \bibinfo{person}{Ziyue Xu}, {and} \bibinfo{person}{Zibin Zheng}.} \bibinfo{year}{2021}\natexlab{}.
\newblock \showarticletitle{FedGL: federated graph learning framework with global self-supervision}.
\newblock \bibinfo{journal}{{\em arXiv preprint arXiv:2105.03170\/}} (\bibinfo{year}{2021}).
\newblock


\bibitem[\protect\citeauthoryear{Clegg, President, Affairs, and Newstead}{Clegg et~al\mbox{.}}{[n. d.]}]%
        {facebookdataresidency}
\bibfield{author}{\bibinfo{person}{Nick Clegg}, \bibinfo{person}{President}, \bibinfo{person}{Global Affairs}, {and} \bibinfo{person}{Jennifer Newstead}.} \bibinfo{year}{[n. d.]}\natexlab{}.
\newblock \bibinfo{title}{Our Response to the Decision on Facebook’s EU-US Data Transfers}.
\newblock   (\bibinfo{year}{[n. d.]}).
\newblock
\newblock
\shownote{\url{https://about.fb.com/news/2023/05/our-response-to-the-decision-on-facebooks-eu-us-data-transfers/}.}


\bibitem[\protect\citeauthoryear{Dwork, Roth, et~al\mbox{.}}{Dwork et~al\mbox{.}}{2014}]%
        {privacybook}
\bibfield{author}{\bibinfo{person}{Cynthia Dwork}, \bibinfo{person}{Aaron Roth}, {et~al\mbox{.}}} \bibinfo{year}{2014}\natexlab{}.
\newblock \showarticletitle{The algorithmic foundations of differential privacy}.
\newblock \bibinfo{journal}{{\em Foundations and Trends{\textregistered} in Theoretical Computer Science\/}} (\bibinfo{year}{2014}).
\newblock


\bibitem[\protect\citeauthoryear{Fang, Cao, Jia, and Gong}{Fang et~al\mbox{.}}{2020}]%
        {fang2020local}
\bibfield{author}{\bibinfo{person}{Minghong Fang}, \bibinfo{person}{Xiaoyu Cao}, \bibinfo{person}{Jinyuan Jia}, {and} \bibinfo{person}{Neil~Zhenqiang Gong}.} \bibinfo{year}{2020}\natexlab{}.
\newblock \showarticletitle{Local model poisoning attacks to byzantine-robust federated learning}. In \bibinfo{booktitle}{{\em Proceedings of the 29th USENIX Conference on Security Symposium}}.
\newblock


\bibitem[\protect\citeauthoryear{Fortunato}{Fortunato}{2010}]%
        {fortunato2010community}
\bibfield{author}{\bibinfo{person}{Santo Fortunato}.} \bibinfo{year}{2010}\natexlab{}.
\newblock \showarticletitle{Community detection in graphs}.
\newblock \bibinfo{journal}{{\em Physics reports\/}} (\bibinfo{year}{2010}).
\newblock


\bibitem[\protect\citeauthoryear{Gandhi and Iyer}{Gandhi and Iyer}{2021}]%
        {gandhi2021p3}
\bibfield{author}{\bibinfo{person}{Swapnil Gandhi} {and} \bibinfo{person}{Anand~Padmanabha Iyer}.} \bibinfo{year}{2021}\natexlab{}.
\newblock \showarticletitle{P3: Distributed deep graph learning at scale}. In \bibinfo{booktitle}{{\em 15th $\{$USENIX$\}$ Symposium on Operating Systems Design and Implementation ($\{$OSDI$\}$ 21)}}. \bibinfo{pages}{551--568}.
\newblock


\bibitem[\protect\citeauthoryear{Geiping, Bauermeister, Dr{\"o}ge, and Moeller}{Geiping et~al\mbox{.}}{2020}]%
        {geiping2020inverting}
\bibfield{author}{\bibinfo{person}{Jonas Geiping}, \bibinfo{person}{Hartmut Bauermeister}, \bibinfo{person}{Hannah Dr{\"o}ge}, {and} \bibinfo{person}{Michael Moeller}.} \bibinfo{year}{2020}\natexlab{}.
\newblock \showarticletitle{Inverting gradients-how easy is it to break privacy in federated learning?}
\newblock \bibinfo{journal}{{\em NeurIPS\/}} (\bibinfo{year}{2020}).
\newblock


\bibitem[\protect\citeauthoryear{Geisler, Schmidt, {\c{S}}irin, Z{\"u}gner, Bojchevski, and G{\"u}nnemann}{Geisler et~al\mbox{.}}{2021}]%
        {geisler2021robustness}
\bibfield{author}{\bibinfo{person}{Simon Geisler}, \bibinfo{person}{Tobias Schmidt}, \bibinfo{person}{Hakan {\c{S}}irin}, \bibinfo{person}{Daniel Z{\"u}gner}, \bibinfo{person}{Aleksandar Bojchevski}, {and} \bibinfo{person}{Stephan G{\"u}nnemann}.} \bibinfo{year}{2021}\natexlab{}.
\newblock \showarticletitle{Robustness of graph neural networks at scale}.
\newblock \bibinfo{journal}{{\em NeurIPS\/}} (\bibinfo{year}{2021}).
\newblock


\bibitem[\protect\citeauthoryear{Hamilton, Ying, and Leskovec}{Hamilton et~al\mbox{.}}{2017}]%
        {hamilton2017inductive}
\bibfield{author}{\bibinfo{person}{Will Hamilton}, \bibinfo{person}{Zhitao Ying}, {and} \bibinfo{person}{Jure Leskovec}.} \bibinfo{year}{2017}\natexlab{}.
\newblock \showarticletitle{Inductive representation learning on large graphs}.
\newblock \bibinfo{journal}{{\em NeurIPS\/}}  \bibinfo{volume}{30} (\bibinfo{year}{2017}).
\newblock


\bibitem[\protect\citeauthoryear{He, Balasubramanian, Ceyani, Rong, Zhao, Huang, Annavaram, and Avestimehr}{He et~al\mbox{.}}{2021}]%
        {he2021fedgraphnn}
\bibfield{author}{\bibinfo{person}{Chaoyang He}, \bibinfo{person}{Keshav Balasubramanian}, \bibinfo{person}{Emir Ceyani}, \bibinfo{person}{Yu Rong}, \bibinfo{person}{Peilin Zhao}, \bibinfo{person}{Junzhou Huang}, \bibinfo{person}{Murali Annavaram}, {and} \bibinfo{person}{Salman Avestimehr}.} \bibinfo{year}{2021}\natexlab{}.
\newblock \showarticletitle{FedGraphNN: {A} Federated Learning System and Benchmark for Graph Neural Networks}.
\newblock \bibinfo{journal}{{\em arXiv preprint arXiv:2104.07145\/}} (\bibinfo{year}{2021}).
\newblock


\bibitem[\protect\citeauthoryear{Hu, Fey, Zitnik, Dong, Ren, Liu, Catasta, and Leskovec}{Hu et~al\mbox{.}}{2020a}]%
        {hu2020open}
\bibfield{author}{\bibinfo{person}{Weihua Hu}, \bibinfo{person}{Matthias Fey}, \bibinfo{person}{Marinka Zitnik}, \bibinfo{person}{Yuxiao Dong}, \bibinfo{person}{Hongyu Ren}, \bibinfo{person}{Bowen Liu}, \bibinfo{person}{Michele Catasta}, {and} \bibinfo{person}{Jure Leskovec}.} \bibinfo{year}{2020}\natexlab{a}.
\newblock \showarticletitle{Open graph benchmark: Datasets for machine learning on graphs}.
\newblock \bibinfo{journal}{{\em NeurIPS\/}} (\bibinfo{year}{2020}).
\newblock


\bibitem[\protect\citeauthoryear{Hu, Ye, Wang, Yu, Zheng, Li, Zhang, Zhang, and Wang}{Hu et~al\mbox{.}}{2020b}]%
        {hu2020featgraph}
\bibfield{author}{\bibinfo{person}{Yuwei Hu}, \bibinfo{person}{Zihao Ye}, \bibinfo{person}{Minjie Wang}, \bibinfo{person}{Jiali Yu}, \bibinfo{person}{Da Zheng}, \bibinfo{person}{Mu Li}, \bibinfo{person}{Zheng Zhang}, \bibinfo{person}{Zhiru Zhang}, {and} \bibinfo{person}{Yida Wang}.} \bibinfo{year}{2020}\natexlab{b}.
\newblock \showarticletitle{Featgraph: A flexible and efficient backend for graph neural network systems}. In \bibinfo{booktitle}{{\em SC20: International Conference for High Performance Computing, Networking, Storage and Analysis}}.
\newblock


\bibitem[\protect\citeauthoryear{Jangda, Polisetty, Guha, and Serafini}{Jangda et~al\mbox{.}}{2021}]%
        {jangda2021accelerating}
\bibfield{author}{\bibinfo{person}{Abhinav Jangda}, \bibinfo{person}{Sandeep Polisetty}, \bibinfo{person}{Arjun Guha}, {and} \bibinfo{person}{Marco Serafini}.} \bibinfo{year}{2021}\natexlab{}.
\newblock \showarticletitle{Accelerating graph sampling for graph machine learning using GPUs}. In \bibinfo{booktitle}{{\em EuroSys}}.
\newblock


\bibitem[\protect\citeauthoryear{Jia, Lin, Gao, Zaharia, and Aiken}{Jia et~al\mbox{.}}{2020}]%
        {jia2020improving}
\bibfield{author}{\bibinfo{person}{Zhihao Jia}, \bibinfo{person}{Sina Lin}, \bibinfo{person}{Mingyu Gao}, \bibinfo{person}{Matei Zaharia}, {and} \bibinfo{person}{Alex Aiken}.} \bibinfo{year}{2020}\natexlab{}.
\newblock \showarticletitle{Improving the accuracy, scalability, and performance of graph neural networks with roc}.
\newblock \bibinfo{journal}{{\em MLSys\/}} (\bibinfo{year}{2020}).
\newblock


\bibitem[\protect\citeauthoryear{Jin, Li, Xu, Wang, Ji, Aggarwal, and Tang}{Jin et~al\mbox{.}}{2021}]%
        {jin2021adversarial}
\bibfield{author}{\bibinfo{person}{Wei Jin}, \bibinfo{person}{Yaxing Li}, \bibinfo{person}{Han Xu}, \bibinfo{person}{Yiqi Wang}, \bibinfo{person}{Shuiwang Ji}, \bibinfo{person}{Charu Aggarwal}, {and} \bibinfo{person}{Jiliang Tang}.} \bibinfo{year}{2021}\natexlab{}.
\newblock \showarticletitle{Adversarial attacks and defenses on graphs}.
\newblock \bibinfo{journal}{{\em ACM SIGKDD Explorations Newsletter\/}} (\bibinfo{year}{2021}).
\newblock


\bibitem[\protect\citeauthoryear{Kairouz, McMahan, Avent, Bellet, Bennis, Bhagoji, Bonawitz, Charles, Cormode, Cummings, et~al\mbox{.}}{Kairouz et~al\mbox{.}}{2021}]%
        {kairouz2021advances}
\bibfield{author}{\bibinfo{person}{Peter Kairouz}, \bibinfo{person}{H~Brendan McMahan}, \bibinfo{person}{Brendan Avent}, \bibinfo{person}{Aur{\'e}lien Bellet}, \bibinfo{person}{Mehdi Bennis}, \bibinfo{person}{Arjun~Nitin Bhagoji}, \bibinfo{person}{Kallista Bonawitz}, \bibinfo{person}{Zachary Charles}, \bibinfo{person}{Graham Cormode}, \bibinfo{person}{Rachel Cummings}, {et~al\mbox{.}}} \bibinfo{year}{2021}\natexlab{}.
\newblock \showarticletitle{Advances and open problems in federated learning}.
\newblock \bibinfo{journal}{{\em Foundations and Trends{\textregistered} in Machine Learning\/}} \bibinfo{volume}{14}, \bibinfo{number}{1--2} (\bibinfo{year}{2021}), \bibinfo{pages}{1--210}.
\newblock


\bibitem[\protect\citeauthoryear{Kipf and Welling}{Kipf and Welling}{2017}]%
        {kipf2016semi}
\bibfield{author}{\bibinfo{person}{Thomas~N Kipf} {and} \bibinfo{person}{Max Welling}.} \bibinfo{year}{2017}\natexlab{}.
\newblock \showarticletitle{Semi-supervised classification with graph convolutional networks}.
\newblock \bibinfo{journal}{{\em ICLR\/}} (\bibinfo{year}{2017}).
\newblock


\bibitem[\protect\citeauthoryear{Kolluri, Baluta, Hooi, and Saxena}{Kolluri et~al\mbox{.}}{2022}]%
        {kolluri2022lpgnet}
\bibfield{author}{\bibinfo{person}{Aashish Kolluri}, \bibinfo{person}{Teodora Baluta}, \bibinfo{person}{Bryan Hooi}, {and} \bibinfo{person}{Prateek Saxena}.} \bibinfo{year}{2022}\natexlab{}.
\newblock \showarticletitle{LPGNet: Link Private Graph Networks for Node Classification}. In \bibinfo{booktitle}{{\em CCS}}.
\newblock


\bibitem[\protect\citeauthoryear{Korolova, Motwani, Nabar, and Xu}{Korolova et~al\mbox{.}}{2008}]%
        {korolova2008link}
\bibfield{author}{\bibinfo{person}{Aleksandra Korolova}, \bibinfo{person}{Rajeev Motwani}, \bibinfo{person}{Shubha~U Nabar}, {and} \bibinfo{person}{Ying Xu}.} \bibinfo{year}{2008}\natexlab{}.
\newblock \showarticletitle{Link privacy in social networks}. In \bibinfo{booktitle}{{\em Proceedings of the 17th ACM conference on Information and knowledge management}}.
\newblock


\bibitem[\protect\citeauthoryear{Li, Tramer, Liang, and Hashimoto}{Li et~al\mbox{.}}{2022}]%
        {li2021large}
\bibfield{author}{\bibinfo{person}{Xuechen Li}, \bibinfo{person}{Florian Tramer}, \bibinfo{person}{Percy Liang}, {and} \bibinfo{person}{Tatsunori Hashimoto}.} \bibinfo{year}{2022}\natexlab{}.
\newblock \showarticletitle{Large language models can be strong differentially private learners}.
\newblock \bibinfo{journal}{{\em ICLR\/}} (\bibinfo{year}{2022}).
\newblock


\bibitem[\protect\citeauthoryear{Lin, Li, Miao, Liu, and Xu}{Lin et~al\mbox{.}}{2020}]%
        {lin2020pagraph}
\bibfield{author}{\bibinfo{person}{Zhiqi Lin}, \bibinfo{person}{Cheng Li}, \bibinfo{person}{Youshan Miao}, \bibinfo{person}{Yunxin Liu}, {and} \bibinfo{person}{Yinlong Xu}.} \bibinfo{year}{2020}\natexlab{}.
\newblock \showarticletitle{Pagraph: Scaling gnn training on large graphs via computation-aware caching}. In \bibinfo{booktitle}{{\em ACM Symposium on Cloud Computing}}.
\newblock


\bibitem[\protect\citeauthoryear{Liu, Wang, and Jiang}{Liu et~al\mbox{.}}{2023b}]%
        {liu2023janus}
\bibfield{author}{\bibinfo{person}{Juncai Liu}, \bibinfo{person}{Jessie~Hui Wang}, {and} \bibinfo{person}{Yimin Jiang}.} \bibinfo{year}{2023}\natexlab{b}.
\newblock \showarticletitle{Janus: A Unified Distributed Training Framework for Sparse Mixture-of-Experts Models}. In \bibinfo{booktitle}{{\em SIGCOMM}}.
\newblock


\bibitem[\protect\citeauthoryear{Liu and Yu}{Liu and Yu}{2022}]%
        {liu2022federated}
\bibfield{author}{\bibinfo{person}{Rui Liu} {and} \bibinfo{person}{Han Yu}.} \bibinfo{year}{2022}\natexlab{}.
\newblock \showarticletitle{Federated Graph Neural Networks: Overview, Techniques and Challenges}.
\newblock \bibinfo{journal}{{\em arXiv preprint arXiv:2202.07256\/}} (\bibinfo{year}{2022}).
\newblock


\bibitem[\protect\citeauthoryear{Liu, Chen, Li, Wu, Zhu, He, Peng, Chen, Chen, and Guo}{Liu et~al\mbox{.}}{2023a}]%
        {liu2023bgl}
\bibfield{author}{\bibinfo{person}{Tianfeng Liu}, \bibinfo{person}{Yangrui Chen}, \bibinfo{person}{Dan Li}, \bibinfo{person}{Chuan Wu}, \bibinfo{person}{Yibo Zhu}, \bibinfo{person}{Jun He}, \bibinfo{person}{Yanghua Peng}, \bibinfo{person}{Hongzheng Chen}, \bibinfo{person}{Hongzhi Chen}, {and} \bibinfo{person}{Chuanxiong Guo}.} \bibinfo{year}{2023}\natexlab{a}.
\newblock \showarticletitle{BGL: GPU-Efficient GNN Training by Optimizing Graph Data I/O and Preprocessing}. In \bibinfo{booktitle}{{\em NSDI}}.
\newblock


\bibitem[\protect\citeauthoryear{Ma, Yang, Miao, Xue, Wu, Zhou, and Dai}{Ma et~al\mbox{.}}{2019}]%
        {ma2019neugraph}
\bibfield{author}{\bibinfo{person}{Lingxiao Ma}, \bibinfo{person}{Zhi Yang}, \bibinfo{person}{Youshan Miao}, \bibinfo{person}{Jilong Xue}, \bibinfo{person}{Ming Wu}, \bibinfo{person}{Lidong Zhou}, {and} \bibinfo{person}{Yafei Dai}.} \bibinfo{year}{2019}\natexlab{}.
\newblock \showarticletitle{{NeuGraph}: Parallel Deep Neural Network Computation on Large Graphs}. In \bibinfo{booktitle}{{\em ATC}}.
\newblock


\bibitem[\protect\citeauthoryear{McMahan, Moore, Ramage, Hampson, and y~Arcas}{McMahan et~al\mbox{.}}{2017}]%
        {mcmahan2017communication}
\bibfield{author}{\bibinfo{person}{Brendan McMahan}, \bibinfo{person}{Eider Moore}, \bibinfo{person}{Daniel Ramage}, \bibinfo{person}{Seth Hampson}, {and} \bibinfo{person}{Blaise~Aguera y Arcas}.} \bibinfo{year}{2017}\natexlab{}.
\newblock \showarticletitle{Communication-efficient learning of deep networks from decentralized data}. In \bibinfo{booktitle}{{\em Artificial intelligence and statistics}}.
\newblock


\bibitem[\protect\citeauthoryear{Md, Misra, Ma, Mohanty, Georganas, Heinecke, Kalamkar, Ahmed, and Avancha}{Md et~al\mbox{.}}{2021}]%
        {md2021distgnn}
\bibfield{author}{\bibinfo{person}{Vasimuddin Md}, \bibinfo{person}{Sanchit Misra}, \bibinfo{person}{Guixiang Ma}, \bibinfo{person}{Ramanarayan Mohanty}, \bibinfo{person}{Evangelos Georganas}, \bibinfo{person}{Alexander Heinecke}, \bibinfo{person}{Dhiraj Kalamkar}, \bibinfo{person}{Nesreen~K Ahmed}, {and} \bibinfo{person}{Sasikanth Avancha}.} \bibinfo{year}{2021}\natexlab{}.
\newblock \showarticletitle{Distgnn: Scalable distributed training for large-scale graph neural networks}. In \bibinfo{booktitle}{{\em Proceedings of the International Conference for High Performance Computing, Networking, Storage and Analysis}}.
\newblock


\bibitem[\protect\citeauthoryear{Mikolov, Chen, Corrado, and Dean}{Mikolov et~al\mbox{.}}{2013}]%
        {mikolov2013efficient}
\bibfield{author}{\bibinfo{person}{Tomas Mikolov}, \bibinfo{person}{Kai Chen}, \bibinfo{person}{Greg Corrado}, {and} \bibinfo{person}{Jeffrey Dean}.} \bibinfo{year}{2013}\natexlab{}.
\newblock \showarticletitle{Efficient estimation of word representations in vector space}.
\newblock \bibinfo{journal}{{\em arXiv preprint arXiv:1301.3781\/}} (\bibinfo{year}{2013}).
\newblock


\bibitem[\protect\citeauthoryear{Moriya, Inoue, Zhang, Olsen, and Skrinak}{Moriya et~al\mbox{.}}{[n. d.]}]%
        {awsdataresidency}
\bibfield{author}{\bibinfo{person}{Hiroki Moriya}, \bibinfo{person}{Yoshitaka Inoue}, \bibinfo{person}{Qiong Zhang}, \bibinfo{person}{Rumi Olsen}, {and} \bibinfo{person}{Kris Skrinak}.} \bibinfo{year}{[n. d.]}\natexlab{}.
\newblock \bibinfo{title}{Building a Cloud-Native Architecture for Vertical Federated Learning on AWS}.
\newblock   (\bibinfo{year}{[n. d.]}).
\newblock
\newblock
\shownote{\url{https://aws.amazon.com/blogs/apn/building-a-cloud-native-architecture-for-vertical-federated-learning-on-aws/}.}


\bibitem[\protect\citeauthoryear{Reisizadeh, Farnia, Pedarsani, and Jadbabaie}{Reisizadeh et~al\mbox{.}}{2020}]%
        {reisizadeh2020robust}
\bibfield{author}{\bibinfo{person}{Amirhossein Reisizadeh}, \bibinfo{person}{Farzan Farnia}, \bibinfo{person}{Ramtin Pedarsani}, {and} \bibinfo{person}{Ali Jadbabaie}.} \bibinfo{year}{2020}\natexlab{}.
\newblock \showarticletitle{Robust federated learning: The case of affine distribution shifts}.
\newblock \bibinfo{journal}{{\em NeurIPS\/}} (\bibinfo{year}{2020}).
\newblock


\bibitem[\protect\citeauthoryear{Roth}{Roth}{[n. d.]}]%
        {metafine}
\bibfield{author}{\bibinfo{person}{Emma Roth}.} \bibinfo{year}{[n. d.]}\natexlab{}.
\newblock \bibinfo{title}{Meta fined 276 million over Facebook data leak involving more than 533 million users}.
\newblock   (\bibinfo{year}{[n. d.]}).
\newblock
\newblock
\shownote{\url{https://www.theverge.com/2022/11/28/23481786/meta-fine-facebook-data-leak-ireland-dpc-gdpr}.}


\bibitem[\protect\citeauthoryear{Rozemberczki, Allen, and Sarkar}{Rozemberczki et~al\mbox{.}}{2021}]%
        {rozemberczki2021multi}
\bibfield{author}{\bibinfo{person}{Benedek Rozemberczki}, \bibinfo{person}{Carl Allen}, {and} \bibinfo{person}{Rik Sarkar}.} \bibinfo{year}{2021}\natexlab{}.
\newblock \showarticletitle{Multi-scale attributed node embedding}.
\newblock \bibinfo{journal}{{\em Journal of Complex Networks\/}} (\bibinfo{year}{2021}).
\newblock


\bibitem[\protect\citeauthoryear{Rozemberczki and Sarkar}{Rozemberczki and Sarkar}{2020}]%
        {rozemberczki2020characteristic}
\bibfield{author}{\bibinfo{person}{Benedek Rozemberczki} {and} \bibinfo{person}{Rik Sarkar}.} \bibinfo{year}{2020}\natexlab{}.
\newblock \showarticletitle{Characteristic functions on graphs: Birds of a feather, from statistical descriptors to parametric models}. In \bibinfo{booktitle}{{\em CIKM}}.
\newblock


\bibitem[\protect\citeauthoryear{Sajadmanesh and Gatica-Perez}{Sajadmanesh and Gatica-Perez}{2021}]%
        {sajadmanesh2021locally}
\bibfield{author}{\bibinfo{person}{Sina Sajadmanesh} {and} \bibinfo{person}{Daniel Gatica-Perez}.} \bibinfo{year}{2021}\natexlab{}.
\newblock \showarticletitle{Locally private graph neural networks}. In \bibinfo{booktitle}{{\em CCS}}.
\newblock


\bibitem[\protect\citeauthoryear{Shen, Jain, and Huang}{Shen et~al\mbox{.}}{2022}]%
        {awsaifinancial}
\bibfield{author}{\bibinfo{person}{Xiaoli Shen}, \bibinfo{person}{Vedant Jain}, {and} \bibinfo{person}{Xin Huang}.} \bibinfo{year}{2022}\natexlab{}.
\newblock \showarticletitle{Detect financial transaction fraud using a Graph Neural Network with Amazon SageMaker}.
\newblock \bibinfo{journal}{{\em AWS Machine Learning Blog\/}} (\bibinfo{year}{2022}).
\newblock
\newblock
\shownote{\url{https://aws.amazon.com/blogs/machine-learning/detect-financial-transaction-fraud-using-a-graph-neural-network-with-amazon-sagemaker}.}


\bibitem[\protect\citeauthoryear{Staff}{Staff}{[n. d.]}]%
        {incountryblog}
\bibfield{author}{\bibinfo{person}{InCountry Staff}.} \bibinfo{year}{[n. d.]}\natexlab{}.
\newblock \bibinfo{title}{Guide to the cross-border transfer of personal data for global companies}.
\newblock   (\bibinfo{year}{[n. d.]}).
\newblock
\newblock
\shownote{\url{https://incountry.com/blog/guide-to-the-cross-border-transfer-of-personal-data-for-global-companies}.}


\bibitem[\protect\citeauthoryear{Thorpe, Qiao, Eyolfson, Teng, Hu, Jia, Wei, Vora, Netravali, Kim, and Xu}{Thorpe et~al\mbox{.}}{2021}]%
        {thorpe21dorylus}
\bibfield{author}{\bibinfo{person}{John Thorpe}, \bibinfo{person}{Yifan Qiao}, \bibinfo{person}{Jonathan Eyolfson}, \bibinfo{person}{Shen Teng}, \bibinfo{person}{Guanzhou Hu}, \bibinfo{person}{Zhihao Jia}, \bibinfo{person}{Jinliang Wei}, \bibinfo{person}{Keval Vora}, \bibinfo{person}{Ravi Netravali}, \bibinfo{person}{Miryung Kim}, {and} \bibinfo{person}{Guoqing~Harry Xu}.} \bibinfo{year}{2021}\natexlab{}.
\newblock \showarticletitle{Dorylus: Affordable, Scalable, and Accurate {GNN} Training with Distributed {CPU} Servers and Serverless Threads}. In \bibinfo{booktitle}{{\em OSDI}}.
\newblock


\bibitem[\protect\citeauthoryear{Tolpegin, Truex, Gursoy, and Liu}{Tolpegin et~al\mbox{.}}{2020}]%
        {tolpegin2020data}
\bibfield{author}{\bibinfo{person}{Vale Tolpegin}, \bibinfo{person}{Stacey Truex}, \bibinfo{person}{Mehmet~Emre Gursoy}, {and} \bibinfo{person}{Ling Liu}.} \bibinfo{year}{2020}\natexlab{}.
\newblock \showarticletitle{Data poisoning attacks against federated learning systems}. In \bibinfo{booktitle}{{\em ESORICS}}.
\newblock


\bibitem[\protect\citeauthoryear{Tramer and Boneh}{Tramer and Boneh}{2021}]%
        {tramer2020differentially}
\bibfield{author}{\bibinfo{person}{Florian Tramer} {and} \bibinfo{person}{Dan Boneh}.} \bibinfo{year}{2021}\natexlab{}.
\newblock \showarticletitle{Differentially private learning needs better features (or much more data)}.
\newblock \bibinfo{journal}{{\em ICLR\/}} (\bibinfo{year}{2021}).
\newblock


\bibitem[\protect\citeauthoryear{Veli{\v{c}}kovi{\'c}, Cucurull, Casanova, Romero, Lio, and Bengio}{Veli{\v{c}}kovi{\'c} et~al\mbox{.}}{2018}]%
        {velivckovic2017graph}
\bibfield{author}{\bibinfo{person}{Petar Veli{\v{c}}kovi{\'c}}, \bibinfo{person}{Guillem Cucurull}, \bibinfo{person}{Arantxa Casanova}, \bibinfo{person}{Adriana Romero}, \bibinfo{person}{Pietro Lio}, {and} \bibinfo{person}{Yoshua Bengio}.} \bibinfo{year}{2018}\natexlab{}.
\newblock \showarticletitle{Graph attention networks}.
\newblock \bibinfo{journal}{{\em ICLR\/}} (\bibinfo{year}{2018}).
\newblock


\bibitem[\protect\citeauthoryear{Wan, Li, Li, Kim, and Lin}{Wan et~al\mbox{.}}{2022}]%
        {wan2022bns}
\bibfield{author}{\bibinfo{person}{Cheng Wan}, \bibinfo{person}{Youjie Li}, \bibinfo{person}{Ang Li}, \bibinfo{person}{Nam~Sung Kim}, {and} \bibinfo{person}{Yingyan Lin}.} \bibinfo{year}{2022}\natexlab{}.
\newblock \showarticletitle{BNS-GCN: Efficient full-graph training of graph convolutional networks with partition-parallelism and random boundary node sampling}.
\newblock \bibinfo{journal}{{\em MLSys\/}} (\bibinfo{year}{2022}).
\newblock


\bibitem[\protect\citeauthoryear{Wang, Li, Li, and Chen}{Wang et~al\mbox{.}}{2022b}]%
        {wang2020graphfl}
\bibfield{author}{\bibinfo{person}{Binghui Wang}, \bibinfo{person}{Ang Li}, \bibinfo{person}{Hai Li}, {and} \bibinfo{person}{Yiran Chen}.} \bibinfo{year}{2022}\natexlab{b}.
\newblock \showarticletitle{Graphfl: A federated learning framework for semi-supervised node classification on graphs}.
\newblock \bibinfo{journal}{{\em ICDM\/}} (\bibinfo{year}{2022}).
\newblock


\bibitem[\protect\citeauthoryear{Wang, Zheng, Ye, Gan, Li, Song, Zhou, Ma, Yu, Gai, Xiao, He, Karypis, Li, and Zhang}{Wang et~al\mbox{.}}{2019}]%
        {wang2019dgl}
\bibfield{author}{\bibinfo{person}{Minjie Wang}, \bibinfo{person}{Da Zheng}, \bibinfo{person}{Zihao Ye}, \bibinfo{person}{Quan Gan}, \bibinfo{person}{Mufei Li}, \bibinfo{person}{Xiang Song}, \bibinfo{person}{Jinjing Zhou}, \bibinfo{person}{Chao Ma}, \bibinfo{person}{Lingfan Yu}, \bibinfo{person}{Yu Gai}, \bibinfo{person}{Tianjun Xiao}, \bibinfo{person}{Tong He}, \bibinfo{person}{George Karypis}, \bibinfo{person}{Jinyang Li}, {and} \bibinfo{person}{Zheng Zhang}.} \bibinfo{year}{2019}\natexlab{}.
\newblock \showarticletitle{Deep Graph Library: A Graph-Centric, Highly-Performant Package for Graph Neural Networks}.
\newblock \bibinfo{journal}{{\em arXiv preprint arXiv:1909.01315\/}} (\bibinfo{year}{2019}).
\newblock


\bibitem[\protect\citeauthoryear{Wang, Kuang, Xie, Yao, Li, Ding, and Zhou}{Wang et~al\mbox{.}}{2022a}]%
        {wang2022federatedscope}
\bibfield{author}{\bibinfo{person}{Zhen Wang}, \bibinfo{person}{Weirui Kuang}, \bibinfo{person}{Yuexiang Xie}, \bibinfo{person}{Liuyi Yao}, \bibinfo{person}{Yaliang Li}, \bibinfo{person}{Bolin Ding}, {and} \bibinfo{person}{Jingren Zhou}.} \bibinfo{year}{2022}\natexlab{a}.
\newblock \showarticletitle{FederatedScope-GNN: Towards a Unified, Comprehensive and Efficient Package for Federated Graph Learning}.
\newblock \bibinfo{journal}{{\em KDD\/}} (\bibinfo{year}{2022}).
\newblock


\bibitem[\protect\citeauthoryear{Wondracek, Holz, Kirda, and Kruegel}{Wondracek et~al\mbox{.}}{2010}]%
        {wondracek2010practical}
\bibfield{author}{\bibinfo{person}{Gilbert Wondracek}, \bibinfo{person}{Thorsten Holz}, \bibinfo{person}{Engin Kirda}, {and} \bibinfo{person}{Christopher Kruegel}.} \bibinfo{year}{2010}\natexlab{}.
\newblock \showarticletitle{A practical attack to de-anonymize social network users}. In \bibinfo{booktitle}{{\em 2010 IEEE symposium on security and privacy}}.
\newblock


\bibitem[\protect\citeauthoryear{Wu, Long, Zhang, and Li}{Wu et~al\mbox{.}}{[n. d.]}]%
        {wu2022linkteller}
\bibfield{author}{\bibinfo{person}{Fan Wu}, \bibinfo{person}{Yunhui Long}, \bibinfo{person}{Ce Zhang}, {and} \bibinfo{person}{Bo Li}.} \bibinfo{year}{[n. d.]}\natexlab{}.
\newblock \showarticletitle{Linkteller: Recovering private edges from graph neural networks via influence analysis}. In \bibinfo{booktitle}{{\em 2022 IEEE Symposium on Security and Privacy (SP)}}.
\newblock


\bibitem[\protect\citeauthoryear{Wu, Pan, Chen, Long, Zhang, and Philip}{Wu et~al\mbox{.}}{2020}]%
        {wu2020comprehensive}
\bibfield{author}{\bibinfo{person}{Zonghan Wu}, \bibinfo{person}{Shirui Pan}, \bibinfo{person}{Fengwen Chen}, \bibinfo{person}{Guodong Long}, \bibinfo{person}{Chengqi Zhang}, {and} \bibinfo{person}{S~Yu Philip}.} \bibinfo{year}{2020}\natexlab{}.
\newblock \showarticletitle{A comprehensive survey on graph neural networks}.
\newblock \bibinfo{journal}{{\em IEEE transactions on neural networks and learning systems\/}} (\bibinfo{year}{2020}).
\newblock


\bibitem[\protect\citeauthoryear{Yang, Chen, and Li}{Yang et~al\mbox{.}}{[n. d.]}]%
        {linkedin}
\bibfield{author}{\bibinfo{person}{Jaewon Yang}, \bibinfo{person}{Jiatong Chen}, {and} \bibinfo{person}{Yanen Li}.} \bibinfo{year}{[n. d.]}\natexlab{}.
\newblock \bibinfo{title}{Completing a member knowledge graph with Graph Neural Networks}.
\newblock   (\bibinfo{year}{[n. d.]}).
\newblock
\newblock
\shownote{\url{https://engineering.linkedin.com/blog/2021/completing-a-member-knowledge-graph-with-graph-neural-networks}.}


\bibitem[\protect\citeauthoryear{Yang, Cohen, and Salakhudinov}{Yang et~al\mbox{.}}{2016}]%
        {yang2016revisiting}
\bibfield{author}{\bibinfo{person}{Zhilin Yang}, \bibinfo{person}{William Cohen}, {and} \bibinfo{person}{Ruslan Salakhudinov}.} \bibinfo{year}{2016}\natexlab{}.
\newblock \showarticletitle{Revisiting semi-supervised learning with graph embeddings}. In \bibinfo{booktitle}{{\em ICML}}.
\newblock


\bibitem[\protect\citeauthoryear{Yin, Mallya, Vahdat, Alvarez, Kautz, and Molchanov}{Yin et~al\mbox{.}}{2021a}]%
        {yin2021see}
\bibfield{author}{\bibinfo{person}{Hongxu Yin}, \bibinfo{person}{Arun Mallya}, \bibinfo{person}{Arash Vahdat}, \bibinfo{person}{Jose~M Alvarez}, \bibinfo{person}{Jan Kautz}, {and} \bibinfo{person}{Pavlo Molchanov}.} \bibinfo{year}{2021}\natexlab{a}.
\newblock \showarticletitle{See through gradients: Image batch recovery via gradinversion}. In \bibinfo{booktitle}{{\em CVPR}}.
\newblock


\bibitem[\protect\citeauthoryear{Yin, Zhu, and Hu}{Yin et~al\mbox{.}}{2021b}]%
        {yin2021comprehensive}
\bibfield{author}{\bibinfo{person}{Xuefei Yin}, \bibinfo{person}{Yanming Zhu}, {and} \bibinfo{person}{Jiankun Hu}.} \bibinfo{year}{2021}\natexlab{b}.
\newblock \showarticletitle{A comprehensive survey of privacy-preserving federated learning: A taxonomy, review, and future directions}.
\newblock \bibinfo{journal}{{\em ACM Computing Surveys (CSUR)\/}} (\bibinfo{year}{2021}).
\newblock


\bibitem[\protect\citeauthoryear{Ying, He, Chen, Eksombatchai, Hamilton, and Leskovec}{Ying et~al\mbox{.}}{2018}]%
        {ying2018graph}
\bibfield{author}{\bibinfo{person}{Rex Ying}, \bibinfo{person}{Ruining He}, \bibinfo{person}{Kaifeng Chen}, \bibinfo{person}{Pong Eksombatchai}, \bibinfo{person}{William~L Hamilton}, {and} \bibinfo{person}{Jure Leskovec}.} \bibinfo{year}{2018}\natexlab{}.
\newblock \showarticletitle{Graph convolutional neural networks for web-scale recommender systems}. In \bibinfo{booktitle}{{\em Proceedings of the 24th ACM SIGKDD international conference on knowledge discovery \& data mining}}.
\newblock


\bibitem[\protect\citeauthoryear{Zhang, Menon, Veit, Bhojanapalli, Kumar, and Sra}{Zhang et~al\mbox{.}}{2021a}]%
        {zhang2021coping}
\bibfield{author}{\bibinfo{person}{Jingzhao Zhang}, \bibinfo{person}{Aditya~Krishna Menon}, \bibinfo{person}{Andreas Veit}, \bibinfo{person}{Srinadh Bhojanapalli}, \bibinfo{person}{Sanjiv Kumar}, {and} \bibinfo{person}{Suvrit Sra}.} \bibinfo{year}{2021}\natexlab{a}.
\newblock \showarticletitle{Coping with label shift via distributionally robust optimisation}.
\newblock \bibinfo{journal}{{\em ICLR\/}}.
\newblock


\bibitem[\protect\citeauthoryear{Zhang, Yang, Li, Sun, and Yiu}{Zhang et~al\mbox{.}}{2021b}]%
        {zhang2021subgraph}
\bibfield{author}{\bibinfo{person}{Ke Zhang}, \bibinfo{person}{Carl Yang}, \bibinfo{person}{Xiaoxiao Li}, \bibinfo{person}{Lichao Sun}, {and} \bibinfo{person}{Siu~Ming Yiu}.} \bibinfo{year}{2021}\natexlab{b}.
\newblock \showarticletitle{Subgraph federated learning with missing neighbor generation}.
\newblock \bibinfo{journal}{{\em NeurIPS\/}} (\bibinfo{year}{2021}).
\newblock


\bibitem[\protect\citeauthoryear{Zheng, Chen, Cheng, Song, Wu, Li, Cheng, Yang, and Zhang}{Zheng et~al\mbox{.}}{2022}]%
        {zheng2022bytegnn}
\bibfield{author}{\bibinfo{person}{Chenguang Zheng}, \bibinfo{person}{Hongzhi Chen}, \bibinfo{person}{Yuxuan Cheng}, \bibinfo{person}{Zhezheng Song}, \bibinfo{person}{Yifan Wu}, \bibinfo{person}{Changji Li}, \bibinfo{person}{James Cheng}, \bibinfo{person}{Hao Yang}, {and} \bibinfo{person}{Shuai Zhang}.} \bibinfo{year}{2022}\natexlab{}.
\newblock \showarticletitle{ByteGNN: efficient graph neural network training at large scale}.
\newblock \bibinfo{journal}{{\em VLDB\/}} (\bibinfo{year}{2022}).
\newblock


\bibitem[\protect\citeauthoryear{Zheng, Ma, Wang, Zhou, Su, Song, Gan, Zhang, and Karypis}{Zheng et~al\mbox{.}}{2020}]%
        {zheng2020distdgl}
\bibfield{author}{\bibinfo{person}{Da Zheng}, \bibinfo{person}{Chao Ma}, \bibinfo{person}{Minjie Wang}, \bibinfo{person}{Jinjing Zhou}, \bibinfo{person}{Qidong Su}, \bibinfo{person}{Xiang Song}, \bibinfo{person}{Quan Gan}, \bibinfo{person}{Zheng Zhang}, {and} \bibinfo{person}{George Karypis}.} \bibinfo{year}{2020}\natexlab{}.
\newblock \showarticletitle{Distdgl: distributed graph neural network training for billion-scale graphs}. In \bibinfo{booktitle}{{\em 2020 IEEE/ACM 10th Workshop on Irregular Applications: Architectures and Algorithms (IA3)}}.
\newblock


\bibitem[\protect\citeauthoryear{Zheng and Saupe}{Zheng and Saupe}{2023}]%
        {amazonscalable}
\bibfield{author}{\bibinfo{person}{Da Zheng} {and} \bibinfo{person}{Florian Saupe}.} \bibinfo{year}{2023}\natexlab{}.
\newblock \showarticletitle{Fast-track graph ML with GraphStorm: A new way to solve problems on enterprise-scale graphs}.
\newblock \bibinfo{journal}{{\em AWS Machine Learning Blog\/}} (\bibinfo{year}{2023}).
\newblock
\newblock
\shownote{\url{https://aws.amazon.com/blogs/machine-learning/fast-track-graph-ml-with-graphstorm-a-new-way-to-solve-problems-on-enterprise-scale-graphs/}.}


\bibitem[\protect\citeauthoryear{Zheng, Zhou, Chen, Wu, Wang, and Zhang}{Zheng et~al\mbox{.}}{2021}]%
        {zheng2021asfgnn}
\bibfield{author}{\bibinfo{person}{Longfei Zheng}, \bibinfo{person}{Jun Zhou}, \bibinfo{person}{Chaochao Chen}, \bibinfo{person}{Bingzhe Wu}, \bibinfo{person}{Li Wang}, {and} \bibinfo{person}{Benyu Zhang}.} \bibinfo{year}{2021}\natexlab{}.
\newblock \showarticletitle{ASFGNN: Automated separated-federated graph neural network}.
\newblock \bibinfo{journal}{{\em Peer-to-Peer Networking and Applications\/}} (\bibinfo{year}{2021}).
\newblock


\bibitem[\protect\citeauthoryear{Zhou, Cui, Hu, Zhang, Yang, Liu, Wang, Li, and Sun}{Zhou et~al\mbox{.}}{2020}]%
        {zhou2020graph}
\bibfield{author}{\bibinfo{person}{Jie Zhou}, \bibinfo{person}{Ganqu Cui}, \bibinfo{person}{Shengding Hu}, \bibinfo{person}{Zhengyan Zhang}, \bibinfo{person}{Cheng Yang}, \bibinfo{person}{Zhiyuan Liu}, \bibinfo{person}{Lifeng Wang}, \bibinfo{person}{Changcheng Li}, {and} \bibinfo{person}{Maosong Sun}.} \bibinfo{year}{2020}\natexlab{}.
\newblock \showarticletitle{Graph neural networks: A review of methods and applications}.
\newblock \bibinfo{journal}{{\em AI Open\/}} (\bibinfo{year}{2020}).
\newblock


\bibitem[\protect\citeauthoryear{Zhu, Liu, and Han}{Zhu et~al\mbox{.}}{2019a}]%
        {zhu2019deep}
\bibfield{author}{\bibinfo{person}{Ligeng Zhu}, \bibinfo{person}{Zhijian Liu}, {and} \bibinfo{person}{Song Han}.} \bibinfo{year}{2019}\natexlab{a}.
\newblock \showarticletitle{Deep leakage from gradients}.
\newblock \bibinfo{journal}{{\em NeurIPS\/}} (\bibinfo{year}{2019}).
\newblock


\bibitem[\protect\citeauthoryear{Zhu, Ponomareva, Han, and Perozzi}{Zhu et~al\mbox{.}}{2021}]%
        {zhu2021shift}
\bibfield{author}{\bibinfo{person}{Qi Zhu}, \bibinfo{person}{Natalia Ponomareva}, \bibinfo{person}{Jiawei Han}, {and} \bibinfo{person}{Bryan Perozzi}.} \bibinfo{year}{2021}\natexlab{}.
\newblock \showarticletitle{Shift-robust gnns: Overcoming the limitations of localized graph training data}.
\newblock \bibinfo{journal}{{\em NeurIPS\/}} (\bibinfo{year}{2021}).
\newblock


\bibitem[\protect\citeauthoryear{Zhu, Zhao, Yang, Lin, Zhou, Ai, Li, and Zhou}{Zhu et~al\mbox{.}}{2019b}]%
        {zhu2019aligraph}
\bibfield{author}{\bibinfo{person}{Rong Zhu}, \bibinfo{person}{Kun Zhao}, \bibinfo{person}{Hongxia Yang}, \bibinfo{person}{Wei Lin}, \bibinfo{person}{Chang Zhou}, \bibinfo{person}{Baole Ai}, \bibinfo{person}{Yong Li}, {and} \bibinfo{person}{Jingren Zhou}.} \bibinfo{year}{2019}\natexlab{b}.
\newblock \showarticletitle{Aligraph: A comprehensive graph neural network platform}.
\newblock \bibinfo{journal}{{\em VLDB\/}} (\bibinfo{year}{2019}).
\newblock


\bibitem[\protect\citeauthoryear{Zhu, Tan, and Xiao}{Zhu et~al\mbox{.}}{2023}]%
        {zhu2023blink}
\bibfield{author}{\bibinfo{person}{Xiaochen Zhu}, \bibinfo{person}{Vincent~YF Tan}, {and} \bibinfo{person}{Xiaokui Xiao}.} \bibinfo{year}{2023}\natexlab{}.
\newblock \showarticletitle{Blink: Link Local Differential Privacy in Graph Neural Networks via Bayesian Estimation}. In \bibinfo{booktitle}{{\em CCS}}.
\newblock


\bibitem[\protect\citeauthoryear{Z{\"u}gner, Borchert, Akbarnejad, and G{\"u}nnemann}{Z{\"u}gner et~al\mbox{.}}{2020}]%
        {zugner2020adversarial}
\bibfield{author}{\bibinfo{person}{Daniel Z{\"u}gner}, \bibinfo{person}{Oliver Borchert}, \bibinfo{person}{Amir Akbarnejad}, {and} \bibinfo{person}{Stephan G{\"u}nnemann}.} \bibinfo{year}{2020}\natexlab{}.
\newblock \showarticletitle{Adversarial attacks on graph neural networks: Perturbations and their patterns}.
\newblock \bibinfo{journal}{{\em ACM Transactions on Knowledge Discovery from Data (TKDD)\/}} (\bibinfo{year}{2020}).
\newblock


\end{thebibliography}
\bibliographystyle{ACM-Reference-Format}


\newpage

\appendix


We include additional information that we have omitted in our main paper in the following Appendix.

\section{Training GNNs with \tool}
\subsection{Communication-efficiency Analysis}
\label{appdx:commanalysis}

Here we describe how we arrive at the equations in Section~\ref{sec:commanalysismain}. Consider a graph randomly split between $n$ workers and each worker has $B$ boundary nodes spread across all other workers. The workers and master collaborate to train a $\num$-layer GNN. Each layer $\hopiter$ has size $L_{\hopiter}$ which outputs an intermediate representation of size $H_\hopiter$. Note that the sum of all layer sizes is the model size $M$. The feature size is $H_0$ and the output embedding size is $H_{\num}$. Further, to train each layer except the last one the lazy-message passing procedure adds a new classifier function to the layer. The size of the classifier function would be $C_{\hopiter}=H_{\hopiter}\times H_{\num}$ since the classifier function takes the representation as input and outputs a vector of size of the last layer's embedding.  Hence, the total layer size to train becomes $L_{\hopiter} + C_{\hopiter}$.   Consider that the model takes $R$ rounds to converge using the standard training. Correspondingly, using the lazy message passing procedure each layer of the GNN is independently trained for $R$ training rounds. We do this to make the comparison fair to the standard training. Also, in this way, each layer of the GNN is trained for $R$ rounds in both the standard and lazy message passing procedures.

\paragraph{Local training phase.} Lazy message passing only requires $\num$-message passing rounds, one before training each layer. In each message passing round, the workers exchange representations of the boundary nodes with each other. Specifically, each worker receives the representations of their boundary nodes from other workers and sends the representations of its inner nodes that are boundary nodes for other workers. If a worker requests representations of $B$ boundary nodes then on average it will also send representations of its inner nodes $B$ times (since the nodes are randomly partitioned). The representations have size $H_\hopiter$. Therefore, the total data volume communicated, $DV^{lazy}_{lt}$ among the $n$ workers is as follows.

\begin{align*}
    DV^{lazy}_{lt} &= 2n\cdot B\cdot\sum_{\hopiter=1}^{\num-1}H_{\hopiter} + 2n\cdot B \cdot H_0
\end{align*}

Now, consider the standard training procedure. Recall that in each training round, there are $2\num$ message passing rounds equally split between forward and backward passes. During the forward pass, the workers send and receive representations of the boundary nodes. During the backward pass, they send and receive the gradients corresponding to the previously shared representations. For instance, if a ``target'' worker receives $H_1$ of a boundary node from a ``source'' worker during the forward pass then the target worker will send the gradient of $H_1$ to that source worker. Therefore the size of the exchanged gradients is the same as the size of the corresponding representations. Across all training rounds, the features, $H_0$ are the only representations that do not change and do not produce gradients since they are non-trainable fixed vectors. Therefore, once the features of the boundary nodes are shared in one training round they can be cached for future training rounds too. Therefore, after caching, across $R$ training rounds the total data volume,  $DV^{standard}_{lt}$, communicated is as follows.

\begin{align*}
    DV^{standard}_{lt} &= 2n\cdot R\cdot B\cdot (\sum_{\hopiter=1}^{\num-1}{H_\hopiter} + \sum_{\hopiter=1}^{\num-1}{H_\hopiter}) + 2n\cdot B \cdot H_0
\end{align*}

Therefore, the ratio can be computed as, 

\begin{align*}
    \frac{DV^{standard}_{lt}}{DV^{lazy}_{lt}} &= \frac{4n\cdot R\cdot B\cdot\sum_{\hopiter=1}^{\num-1}{H_\hopiter + 2n\cdot B \cdot H_0}}{2n\cdot B\cdot\sum_{\hopiter=1}^{\num-1}H_{\hopiter} + 2n\cdot B \cdot H_0}\\
    &=\frac{2R\cdot\sum_{\hopiter=1}^{\num-1}H_{\hopiter} + H_0}{\sum_{\hopiter=1}^{\num-1}H_{\hopiter} + H_0}
\end{align*}

Therefore, if $H_0 \leq H_{\hopiter}$, the data volume is approximately $\Theta(R)\times$ higher for standard than lazy message passing. $R$ can be arbitrary in practice depending on the specific application scenario. GNNs for node classification have been trained even for $3000$ rounds in prior works~\cite{wan2022bns}.

\paragraph{Gradient aggregation phase.} In every training round, each worker in the standard training procedure sends their gradient vector of model size $M$ to the master for aggregation. Post aggregation, each client receives the updated model of size $M$ to start the next local training phase. Therefore, the total network data volume, $DV^{standard}_{reduce}$, communicated by all clients across $R$ training rounds during this phase is as follows.

\begin{align*}
    DV^{standard}_{reduce} = n\cdot R\cdot M + n\cdot R \cdot M
\end{align*}

For the lazy message passing procedure, the gradient aggregation phase occurs while training for each GNN layer in a similar fashion. Therefore, if the layer sizes are $L_{\hopiter} + C_{\hopiter}$ then the total network data volume, communicated is as follows.

\begin{align*}
    DV^{lazy}_{reduce} &= 2n\cdot R\cdot(\sum_{\hopiter=1}^{\num}{L_\hopiter} + \sum_{\hopiter=1}^{\num-1}{C_{\hopiter}})\\
    &= 2n\cdot R\cdot M + 2n\cdot R\cdot\sum_{\hopiter=1}^{\num-1}{C_{\hopiter}}
\end{align*}

Therefore, the ratio can be computed as,

\begin{align*}
    \frac{DV^{standard}_{reduce}}{DV^{lazy}_{reduce}} &= \frac{2n\cdot R\cdot M}{2n\cdot R\cdot M + 2n\cdot R\cdot\sum_{\hopiter=1}^{\num-1}{C_{\hopiter}}}\\
    &= \frac{1}{1 + \frac{\sum_{\hopiter=1}^{\num-1}{C_{\hopiter}}}{M}}
\end{align*}

The ratio is close to $1$ as the model size $M$ is usually much larger than the total size of additional classifier layers added. For the GNNs we evaluated, we find that the model size $M$ can be more than $20\times$   the sum of the sizes of classifier layers added. Therefore, in that scenario, the ratio would be $\frac{1}{1+0.05}=0.95$. Effectively, there may not be much difference between the data volumes communicated during this phase even though the GNN is being trained layer-by-layer for $R$ rounds each with additional classifier layers.

Following from the aforementioned analysis, we use eq. ~\ref{eq:2} to compute the theoretically expected $\gamma$, representing the ratio of network data volumes transferred during standard training versus lazy. Subsequently, we compare this theoretical value with the empirically observed ratios obtained from our experiments in table \ref{tab:gamma_comp_products} and \ref{tab:gamma_comp_reddit}.

\begin{table}
    \centering
    \caption{Theoretically expected versus empirically observed values of $\gamma$ for \products}
    \begin{tabular}{c|c|c}
        \#Paritions &  Theoretical & Empirical \\\hline
        $2$ & $314.2$ & $314.6$ \\
        $4$ & $319.8$ & $320.2$ \\
        $6$ & $320.5$  & $321.2$ \\
        $8$ & $320.5$ & $321.1$ \\\hline
    \end{tabular}
    \label{tab:gamma_comp_products}
\end{table}

\begin{table}
    \centering
    \caption{Theoretically expected versus empirically observed values of $\gamma$ for \reddit}
    \begin{tabular}{c|c|c}
        \#Paritions &  Theoretical & Empirical \\\hline
        $2$ & $125.9$ & $125.9$ \\
        $4$ & $146.8$ & $146.8$ \\
        $6$ & $151.5$  & $151.4$ \\
        $8$ & $153.4$ & $153.5$ \\
        $16$ & $155.1$ & $155.2$ \\
        $24$ & $154.7$ & $154.8$ \\\hline
    \end{tabular}
    \label{tab:gamma_comp_reddit}
\end{table}
\section{Evaluation}
\label{appdx:eval}

\begin{table}[ht]
     \centering
    \caption{All datasets' statistics.}
    \resizebox{\columnwidth}{!}{%
    \begin{tabular}{c c c c c}
        \toprule
        Dataset & Nodes & Edges  & Features & Classes\\ \hline
        \cora~\cite{yang2016revisiting}   & 2,708 & 10,566 & 1,433 & 7 \\
        \seer~\cite{yang2016revisiting}   & 3,327 & 9,104 &  3,703 & 6 \\
        \lastfm~\cite{rozemberczki2020characteristic} & 7,624  &  55,612 &  128    & 16 \\
        \pubmed~\cite{yang2016revisiting} & 19,717 & 88,648 & 500 & 3 \\
        \fbpage~\cite{rozemberczki2021multi} & 22,470 & 	342,004 & 128 & 4 \\
        \reddit~\cite{hamilton2017inductive} & 233K & 114M & 602 & 41 \\
        \products~\cite{hu2020open} & 2.4M & 62M & 100 & 47 \\\hline
         
    \end{tabular}
    }
    \label{tab:datastats}
\end{table}

\subsection{Evaluation Setup}
\label{appdx:evalsetup}

Here we provide additional details on our evaluation setup.

\paragraph{Datasets \& Splits.}
\cora, \seer, and \pubmed are some of the first benchmarks to measure the node classification performance of GNNs. These are citation networks where nodes are documents and their features represent the existence/non-existence of certain keywords. \lastfm, \fbpage, and \reddit datasets are taken from their respective social networking platforms. In the \lastfm dataset nodes represent users from Asia and the edges represent friendship. \fbpage is a social network graph where nodes are pages on Facebook and edges represent mutual follower relationships. \reddit dataset consists of posts obtained from the Reddit platform. Two posts have an edge between them if the same users have commented on both posts. \products dataset is taken from Amazon listed products where the nodes are the products, their descriptions are the features, and edges represent co-purchases. In all of the above datasets, the task is to classify the nodes into known set of classes. For instance, in the \products dataset, the classes correspond to top-level product categories on Amazon.

We perform node classification in the transductive setting for all datasets where the graph is fixed and not changing between the training, validation, and testing phases. For all datasets except \reddit and \products we use $20\%$ of the data for training, $10\%$ for validation, and the rest for testing. For \reddit we use the publicly available splits of $66\%$ for training, $10\%$ for validation, and $24\%$ for testing~\cite{hamilton2017inductive}. Similarly, for \products we use the publicly available splits of $8\%$ for training, $2\%$ for validation, and the remaining for testing~\cite{hu2020open}.

\paragraph{Hyperparameters.}
To make our results easily reproducible, we do not tune the hyperparameters. For \reddit and \products benchmarks we follow the same setup as prior work and fully reproduce their reported accuracy~\cite{wan2022bns}. 
\begin{itemize}
    \item For \reddit, we use $4$-layer models for \gcn and \sage based models with learning rate $0.01$ and dropout is $0.5$. For \gat based models, we use $2$ layers since we observe that \gat performs much better with $2$ layers than $4$ and we use a learning rate of $0.003$ with no dropout. 
    \item For \products, we use $3$-layer models for all GNNs. We set the learning rate to $0.003$ with dropout $0.3$ for \gcn and \sage whereas no dropout for \gat.
\end{itemize}

For other datasets, we use GNNs with $2$ aggregation layers and set the learning rate at $0.003$ with a dropout $0.3$. The hidden size for all GNNs is set as $256$ for all datasets. These values for the number of aggregation layers and hidden size are commonly used by the prior work including BNS-GCN~\cite{wan2022bns,hamilton2017inductive,kipf2016semi}. Note that, we use the same hyperparameters used for the base GNN for corresponding \toolgnn and the BNS-GNN. For instance, the hyperparameters used for \gcn will be used for \toolgcn and \gcna too.

\subsection{Feasibility Study: Fully-distributed Setup}
\label{appdx:feasibility}

We implement and deploy end-to-end training of \toolgnns on networked embedded system workers. The fully-distributed setup is motivated by mobile applications where typically there is only one graph node per worker such as a user of a social network or contact network. We cannot demonstrate the fully-distributed setup for even smaller benchmarks such as the \cora dataset since it would require over $2700$ workers. Instead, we demonstrate the training on $2$ devices with the dataset horizontally split between them. We believe that our demonstration has sufficient evidence to show that \toolgnns can be immediately deployed in the mobile and asynchronous fully-distributed setups such as individual users with their mobile phones as workers.

\paragraph{Hardware.} We choose $2$ Raspberry Pi model 4B boards with WiFi and Bluetooth support. One has $2$ GB and another has $4$ GB RAM with no GPU support. We implement a central server (master) on a Ubuntu 20.04 machine with $128$ GB RAM. The server's location is immaterial to show the feasibility of training \toolgnns on workers, however, it may slightly affect the total training time due to network latency. The server is connected to Ethernet with approximately $100$ Mbps upload and download speeds measured at any given random time. The workers communicate with the server over the Internet. The workers communicate with each other over Bluetooth when in proximity. The RPi model 4B has Bluetooth 5.0 version.

\paragraph{Software stack.} We implement \tool's network backend component (see \tool's components in Section~\ref{sec:retexo-arch}) using the Flower library~\cite{beutel2020flower}. The library provides communication API for the worker-to-master communication channels. We build a custom worker-to-worker communication channel using Python socket API on top of the Bluedot library~\footnote{\url{https://github.com/martinohanlon/BlueDot}}. Correspondingly, we build an RSU component and LMP trainer on top of this network backend. The popular libraries used for GNNs such as torch-geometric and torch-scatter do not support the ARM architectures. Therefore, none of the preexisting libraries for GNN training can be used for Raspberry Pis. Therefore, we build custom \toolgnns on just plain PyTorch without using any of the aforementioned libraries.

\paragraph{Dataset, Model, and Split.} We use the \cora dataset and \toolsage with $2$ aggregation layers. We take half of the nodes randomly and store them along with their induced subgraph on one worker. We store the remaining nodes and their induced subgraph of the other worker. On each worker, we also store the boundary nodes and edges. We observe that $1354$ nodes are present on each worker and $1088$ boundary edges. In a fully-distributed setup, there would be a single node in each worker and these boundary edges would just be the single node's neighbors. For train/val/test split we use the publicly available split~\footnote{\url{https://pytorch-geometric.readthedocs.io/en/latest/generated/torch_geometric.datasets.Planetoid.html\#torch_geometric.datasets.Planetoid}}.

\paragraph{Message passing} When the worker-to-worker communication channels are on bluetooth we have to wait until the two RPis come into close proximity. When in proximity, the RPis exchange the embeddings of the necessary cross-worker neighbors. Specifically, each RPi will send a message requesting the other for the embeddings of the nodes corresponding to the cross-worker edges. After receiving the first message, each RPi will send a message consisting of the requested node's embeddings. After the message passing round is finished the RPis will be ready for training the next layer. The message passing round in this setup requires the exchanging of messages between just two workers. In the fully-distributed setup, this round would end after each worker exchanges messages with many other workers. Once the communication channel is available, the time required for exchanging messages between two workers is bounded. However, since the communication channels may not be available, the time required to complete the message passing round can be arbitrary and will depend on many factors such as the end application and the mobility patterns of workers. In practice, a time period can be allocated to complete one message passing round such as $1$ day if it is expected that a worker would be able to exchange embeddings with most/all of its neighboring workers during that time. 

\paragraph{Qualitative Analysis.} We train \toolsage on \cora dataset end-to-end. The manual effort involved in this training is minimal. We manually pair the workers over bluetooth once before starting the training. This is a necessary step but only done once with every neighboring worker. This step also dictates which workers are trusted since only these workers can send and receive data. We manually start the training for each layer in our case study. However, in practice, the server can start the training without any user intervention. For instance, Google does federated learning over mobile devices when they are plugged into charging at night~\cite{bonawitz2019towards}. For message passing, we manually run a script to exchange the required embeddings when the RPis are in proximity. However, this step can also be automated by just making the devices run such a script automatically once they detect that their neighbors are in proximity. Therefore, a user does not have to oversee the training procedure after the initial pairing of the necessary devices required for training. The server can decide how much time to allocate for message passing rounds and after that start a new training round.

\subsection{Limitations \& Future Work}
\label{appdx:limitations}

\paragraph{Trustworthy Distributed Training.} Similar to many prior works and existing real-world deployments~\cite{bonawitz2019towards,he2021fedgraphnn,wang2022federatedscope} we do not design training protocols for \tool with formal privacy guarantees such as differential privacy~\cite{privacybook}. Therefore, training \tool as-is in our setup may leak information about training data via intermediate gradients submitted to the server~\cite{zhu2019deep,geiping2020inverting}. However, we point out that preventing information leakage via gradients is an active area of research (see Survey~\cite{yin2021comprehensive}). Specifically, existing solutions such as secure aggregation that prevent information leakage from individual gradients can be readily integrated with \tool~\cite{bonawitz2017practical}. This, however, does not prevent information leakage from aggregated gradients since the server can see the aggregate gradients~\cite{yin2021see}. Training \toolgnns with differential privacy guarantees can further prevent leakage from aggregated gradients. However, generic differential privacy techniques are notorious for reducing the utility of the models~\cite{tramer2020differentially,li2021large,wu2022linkteller,kolluri2022lpgnet}. Designing differential privacy techniques, for both features and edges, that are tailored to do node classification  in the fully-distributed setup is an interesting future work.

Finally, the training \tool may be susceptible to the robustness issues that are common in distributed setups. For instance, distribution shifts in the features and labels~\cite{reisizadeh2020robust,zhang2021coping,zhu2021shift}, adversarial perturbations to the graph structure~\cite{zugner2020adversarial,bojchevski2019adversarial}, model and data poisoning attacks~\cite{bhagoji2019analyzing, fang2020local,tolpegin2020data}. Integrating existing robust federated learning and graph neural network training techniques~\cite{geisler2021robustness,bojchevski2019certifiable,jin2021adversarial} or designing new ones is an interesting direction for future work.

\paragraph{Feasibility of Training \toolgnns in the Fully-distributed Setup.}  We do not demonstrate training of \toolgnns when only one graph node is available per worker since that would require a large number of individual worker devices. However, we train \toolgnns end-to-end on two devices and claim that training \toolgnns in such setups should be easier than on two devices. Resource-wise our argument is sound since per worker the resources required to train with one node per worker is much smaller than training with the graph data split between just two workers. A worker just has to train a layer based on a single data point which requires little RAM. Also, it has to transmit a single embedding over worker-to-worker communication channels which can be quick ($0.6$ seconds in our case study) even over low bandwidth bluetooth channel (see Section~\ref{sec:evaldecentralized}).

\end{document}